# Advancements in Generative AI: A Comprehensive Review of GANs, GPT, Autoencoders, Diffusion Model, and Transformers.

**Staphord Bengesi[1], Hoda El-Sayed[1], Md Kamruzzaman Sarker[1], Yao Houkpati[1], John Irungu[3] and Timothy Oladunni[2]**

[1] Dept. of Computer Science, Bowie State University, Bowie, MD 20715 USA
[2] Dept. of Computer Science, Morgan State University, Baltimore, MD 21251 USA
[3] Dept. of Computer Science, University of the District of Columbia, Washington, DC 20008 USA

Corresponding author: sbengesi@bowiestate.edu

**ABSTRACT** The launch of ChatGPT has garnered global attention, marking a significant milestone in the field of Generative Artificial Intelligence. While Generative AI has been in effect for the past decade, the introduction of ChatGPT has ignited a new wave of research and innovation in the AI domain. This surge in interest has led to the development and release of numerous cutting-edge tools, such as Bard, Stable Diffusion, DALL-E, Make-A-Video, Runway ML, and Jukebox, among others. These tools exhibit remarkable capabilities, encompassing tasks ranging from text generation and music composition, image creation, video production, code generation, and even scientific work. They are built upon various state-of-the-art models, including Stable Diffusion, transformer models like GPT-3 (recent GPT-4), variational autoencoders, and generative adversarial networks. This advancement in Generative AI presents a wealth of exciting opportunities and, simultaneously, unprecedented challenges. Throughout this paper, we have explored these state-of-the-art models, the diverse array of tasks they can accomplish, the challenges they pose, and the promising future of Generative Artificial Intelligence.

## I. INTRODUCTION

The release of ChatGPT on November 30, 2022 [30][31], triggered an exponential surge in the groundbreaking and widespread popularity of Generative Artificial Intelligence (GAI) to the general public. This remarkable achievement could be traced to the 1956 summer project at Dartmouth College spearheaded by McCarthy; marking the inception of the Artificial Intelligence [1]. The endeavor aimed to develop machines with the ability to perform tasks typically demanding human intelligence [2] [3] [4] [5] [6]. These tasks include computer vision, natural language processing, robotics, and many others. Since then, significant advancements have been achieved in imbuing machines with the capability of talking, walking, thinking, and acting like humans. Notably, a series of algorithms, including the Regression model, perceptron algorithm [7], Decision tree[8], K-Nearest Neighbor [9], Naive Bayes Classifier, Back Propagation, support vector machine (SVM)[10], and Random Forest [11] have emerged. These algorithms in the contemporary are commonly referred to as classical/traditional machine learning algorithms and most of them were developed before the year 2000. Furthermore, there is an advancement in deep learning algorithms, including the development of Convolutional Neural Networks (CNNs) in the 1980s [12], Recurrent Neural Networks (RNNs) in 1985[13], Long Short-Term Memory (LSTM) in 1997 [14], and Bidirectional Long Short-Term Memory (BiLSTM) [15] in the same year. However, until recent times, widespread attention has been limited primarily because of computing resources and dataset availability limitations [16].

To tackle the constraints imposed by limited datasets, researchers from Stanford University, Princeton University, and Columbia University jointly launched the ImageNet Large Scale Visual Recognition Challenge in 2010 [17]. This competition played a pivotal role in driving advancements in neural network architectures, with a particular focus on Convolutional Neural Networks (CNNs). Since then, CNN has been established as algorithm for image classification and computer vision[18]. The breakthrough achievement of AlexNet in 2012 [19] marked a significant milestone in the practical application of deep learning in computer vision tasks. The success of the ImageNet Competition ignited a surge in interest and investment in deep learning research. This newfound enthusiasm resulted in the continuous evolution of improved architectural innovations, including models such as ResNet[20], DenseNet[21], MobileNet[22], and EfficientNet[23]. These models set the gold standard for various cutting-edge technologies, such as transfer learning, continual learning, attention mechanisms [24], self-supervised learning, and generative AI.

Before 2014, all existing deep learning models were primarily descriptive, focusing on summarizing or representing existing data patterns and relationships. These models aimed to explain the data patterns and make predictions based on the information present. However, Goodfellow et al. [25] in 2014 introduced the Generative Adversarial Network (GAN) ushering in a new era of Generative Artificial Intelligence (GAI) realization. Unlike their descriptive counterparts, generative models, such as GANs, are designed to learn the underlying probability distribution of the data [26] . Their primary goal is to generate new data samples that closely resemble the patterns observed in the training data [27][28].

The breakthrough of GAN marked a significant departure from traditional deep learning methods, opening exciting possibilities for Generative artificial intelligence. GAI has since garnered widespread attention due to its transformative impact across various domains of life. It offers elegant solutions to complex problems [29] enabling the creation of synthetic data, artistic content, and realistic simulations. This paradigm shift in AI technology has profoundly influenced the new perception, implementation, and utilization on artificial intelligence, sparking innovation and new application opportunities across industries.

The emergence of GAI has sparked numerous questions, prompting a need for a comprehensive exploration. In that vein, this paper aims at providing an in-depth exploration to the state-of-the-art in GAI, including models, task categorization, applications, areas of influence, challenges, and prospects. To achieve this, our work is structured as follows: Section II introduces contemporary generative models. Section III elaborates on the various tasks within Generative AI. Section IV examines the diverse applications of Generative AI. Section V delves into the outlook for generative AI. Lastly, Section VI offers a conclusion.

## II. GENERATIVE MODELS

There has been a shift in the focus of researchers from discriminative learning to generative learning in the contemporary era. Multiple generative models have emerged with the capability of generating new data points like the training data inputs based on learning their distribution. This section will discuss current state-of-the-art theoretical and mathematical foundations of generative models.

### A. AUTOENCODER

Autoencoder is an unsupervised machine learning neural network model that encodes the input data using an encoder into a lower-dimensional representation (encoding) and then uses a decoder to decode it back to its original form (decoding) while reducing the reconstruction error [32]. This model was primarily designed for Dimensionality Reduction, Feature Extraction, Image Denoising, Image Compression, Image Search, Anomaly Detection and Missing Value Imputation [32].

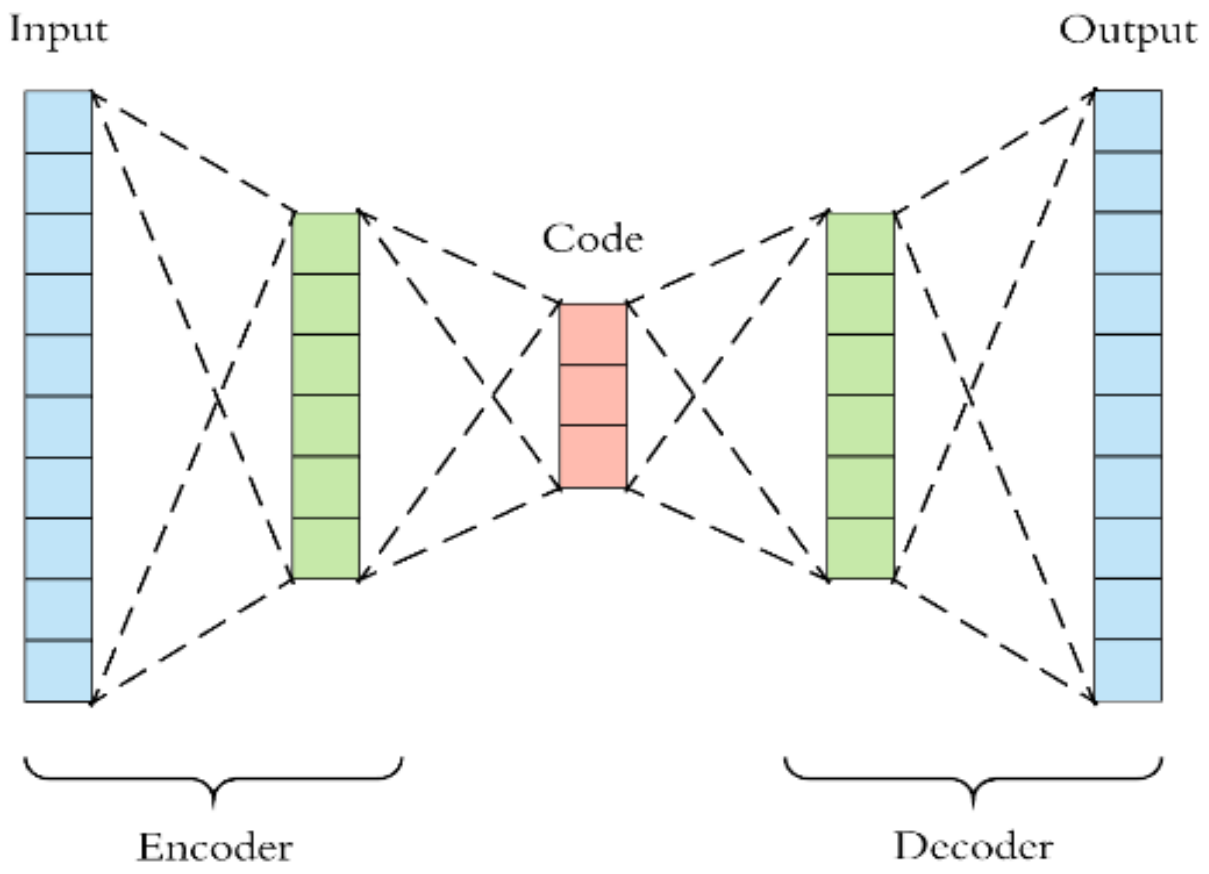

**FIGURE 1.** Autoencoder architecture[1]

Both encoder and decoder of the model are neural networks written as a function of input and a generic function of code layer respectively [33]. Based on figure 1, autoencoder is made up of four components namely:

- **Encoder:** This component reduces and compresses the input data into lower dimensions. As a result of its output, it creates a new layer called code.
- **Code/Bottleneck:** a layer that contains a compressed and the lowest possible dimensions of input data representation.
  Consider equation 1 below.
  $$h_i = f(X_i) \quad (1)$$
  Whereby $\boldsymbol{h_i}$ is code layer after function $\boldsymbol{f}$ *with* user defined parameters is applied to the input $\boldsymbol{X_i}$
- **Decoder**: Reconstructs the code layer from lower dimension representation to input.
- **Reconstruction Loss**: Defines the final output of the decoder, measuring how closely the output resembles the original input.

$$\widetilde{X_i} = g(h_i) \quad (2)$$

[1] Source: https://towardsdatascience.com/applied-deep-learning-part-3-autoencoders-1c083af4d798

Where $\widetilde{X_i}$ is the output of encoder after second generic function to the code layer.

The training of the autoencoder involves minimizing the dissimilarity between the input and the output [33], as shown in Equation 3.

$$Argmin_{f,g} < \Delta(X_i, \widetilde{X_i}) \quad (3)$$

The encoder and the decoder are composed of fully connected feedforward neural networks where the input, code, and output layers consist each of a single neural network layer defined by the user. Like other standard neural networks, autoencoders apply activation functions such as sigmoid and Relu. Various variants of autoencoder exist, such as contractive, Denoising, and sparse autoencoder [34]. Generally, the plain autoencoders prior mentioned are not generative since they do not generate new data but replicate the input. However, the variational autoencoder is the variant that is generative [32].

### 1) VARIATIONAL AUTOENCODER

Variational autoencoder (VAE) evolved as a result of the introduction of variational inference (A statistical technique for approximating complex distributions) to Autoencoder (AE) by Kingma et al. [35]. It's a generative model that utilizes Variational Bayes Inference to describe data generation using a probabilistic distribution [36].

Unlike traditional AEs, VAEs have an extra sampling layer in addition to an encoder and decoder layer as depicted in figure 2. Training the VAEs model involves encoding the input as a distribution over the latent space and generating the latent vector from the distribution sampling. Afterward, the latent vector is decoded, the reconstruction error is computed, and the reconstruction error is backpropagated through the network. During the training process, regularization is introduced explicitly to prevent overfitting.

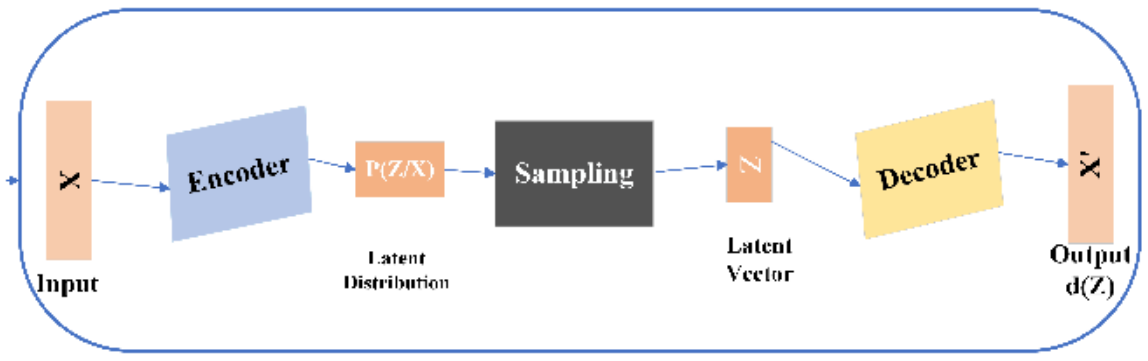

**FIGURE 2.** Variational encoder architecture

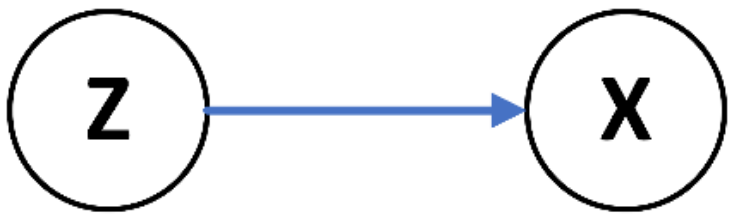

**FIGURE 3.** Probability Model

Probabilistically, VAE is composed of a latent representation ***z*** as depicted by Figure 3, drawn from the prior distribution ***p(z)*** and the data **x** drawn from the conditional likelihood distribution ***p(x|z)*** which is referred to as probabilistic decoder and can be expressed as:

$$p(x,z)=p(x|z)p(z) \quad (4)$$

The inference of the model is examined by computing the posterior of the latent vector using the Bayes theorem shown in equation 5.

$$p(z \mid x) = \frac{p(x|z)p(z)}{p(x)} \quad (5)$$

With any distribution variant such as Gaussian, variational inference can approximate the posterior, and its reliability in approximation can be assessed through Kullback-Leibler divergence which measures the information lost during approximation. This model has significantly influenced generative AI, as demonstrated in Table 1, which highlights a few outstanding state-of-the-art examples using VAE across various domains.

TABLE 1
VAE STATE-OF-THE-ART

| category | Subcategory Domains | Dataset | Reference |
|---|---|---|---|
| Image Processing | Image Classification | MRI datasets, SAR images, ImageNet dataset, NWPU-RESISC45 | [37] [38] [39] [40] |
| | Image Compression | Kodak dataset | [41] |
| | Image Resolution | DIV2K and Flickr2K image dataset | [42] |
| Audio Processing | Noisy voice recorded datasets | NIL | [43] [44] |
| Video Processing | Prediction | MineRL and MMNIST | [45] |
| Video | Infrastructure Monitoring | UCSD Ped2, Fan deterioration simulated dataset | [46] [47] [48] |
| Audio | Infrastructure Monitoring | MIMII | [49] |
| Sensor | Nonlinear analysis | Simulated Dataset, Butane content | [50] [51] |
| | Modeling | DCS | [52] [53] [54] |
| Activity monitoring | Finance | 284,807 credit card transactions | [55] [56] |

### *B. TRANSFORMER*

The ground-breaking work of Vaswani et al. *"Attention Is All You Need"* by the Google Brain team introduced a transformer model which can analyze large-scale dataset [24]. Transform was initially developed for natural language processing (NLP) but was subsequently adapted to other areas of machine learning, such as computer vision [57] [58] [59]. This model aimed to solve RNNs, and CNNs shortcomings such as long-range dependencies, gradient vanishing, gradient explosion, the need for larger training steps to reach a local/global minima, and the fact that parallel computation was not allowed [24]. Thus, the proposed solution presented a novel way of handling neural network tasks like translation, content generation, and sentiment analysis [60]

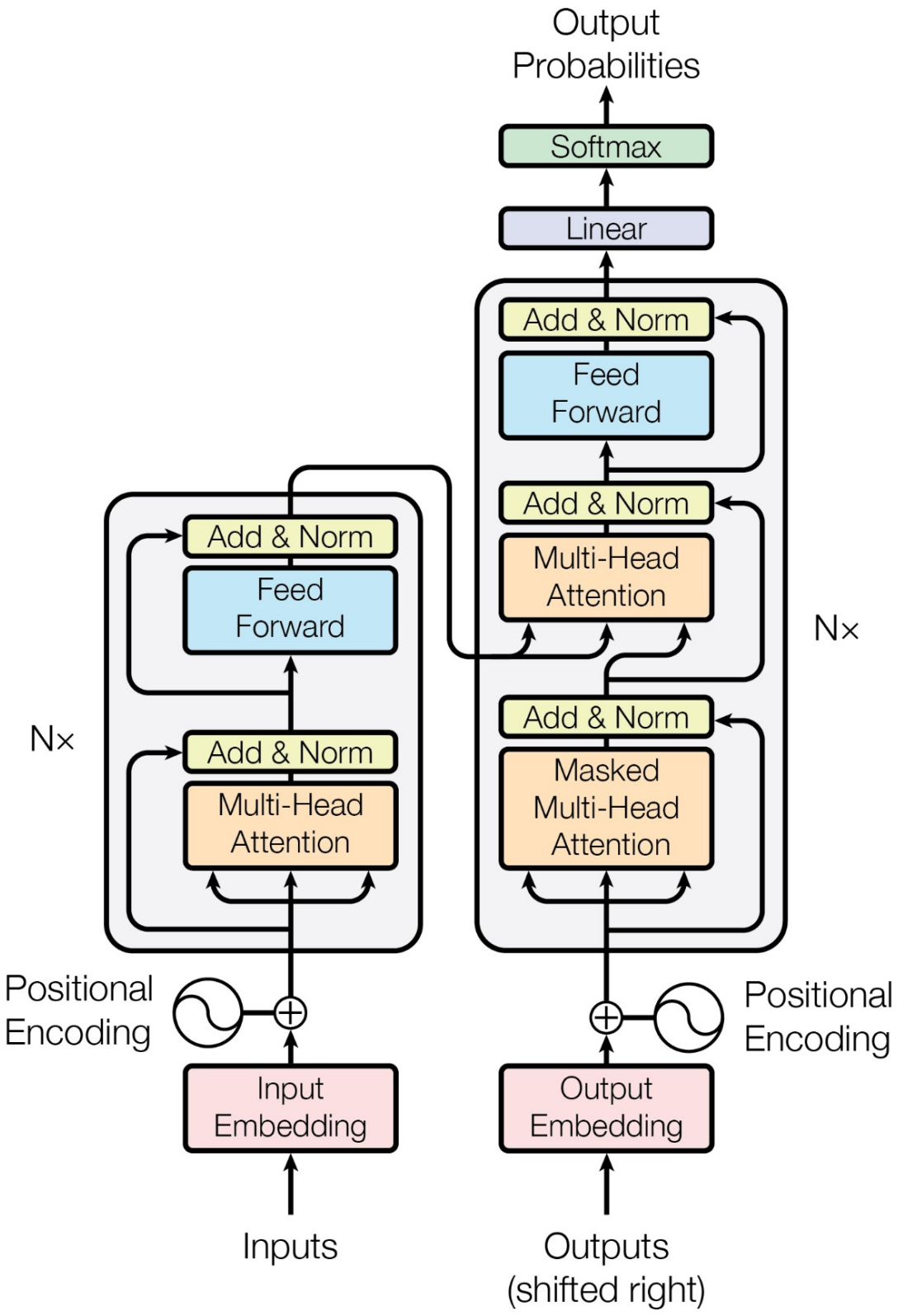

**FIGURE 4.** Transformer Architecture [24]

**Transformer Architecture**

Vaswani et al, introduced three main concepts in their study as depicted in figure 4, including self-attention, which allows a model to evaluate input sequences according to their importance, thus reducing long-range dependencies, multi-head attention which allows the model to learn multiple means of the input sequence, and word embedding, which transforms inputs into vectors.

**Encoder and Decoder**

It is worth mentioning that the transformer architecture (Figure 4) inherits the encoder-decoder structure[61]that utilizes stacked self-attention and point-wise layers, fully connected layers for both the encoder and decoder [62]. The encoder consists of a stack of N = 6 identical layers, each with two sublayers, including a multi-head self-attention mechanism and a fully connected feedforward network. A decoder is like an encoder, but with an additional sublayer which masks the multi-head attention. Encoders and decoders both apply residual connections to the sublayers, followed by normalization of the layers.

**Self-Attention**

Attention describes the mechanism for a better understanding of the word's context by paying attention to the vital part of the sentence or any input. It involves mapping a vector of query and a set of key-value pairs to an output vector. According to [24], self-attention refers to Scaled Dot-Product Attention consisting of queries and key dimensions $d_k$, and dimension $d_v$ values computed according to the following formula:

$$Attention(Q,K,V) = softmax(\frac{QK^T}{\sqrt{d_k}})V \qquad (6)$$

Figure 5 depicts the structure attention whereby the SoftMax activation function is used to compute the weights on values.

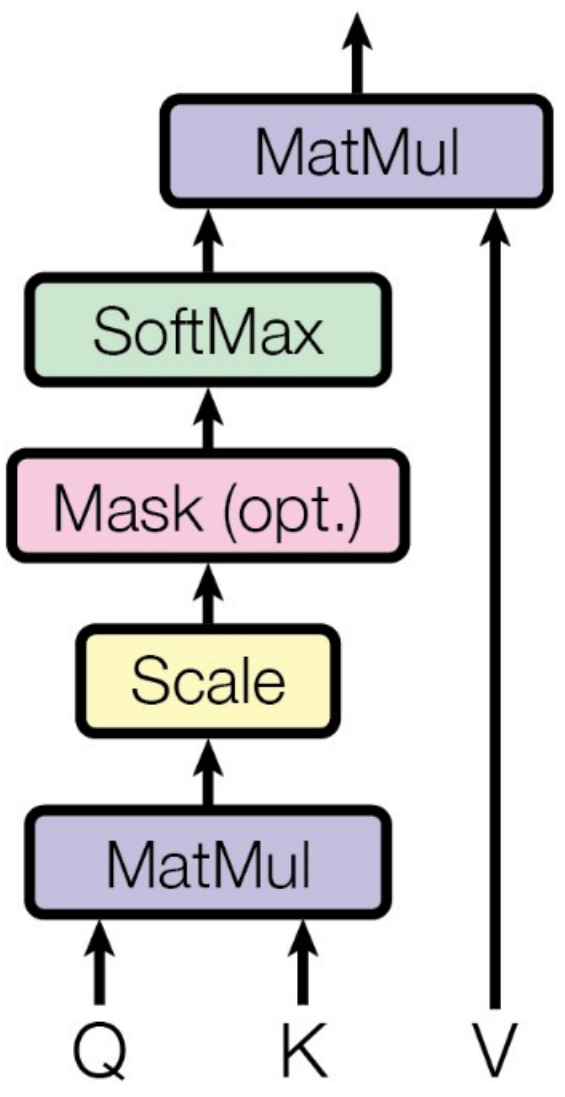

**FIGURE 5.** Self-attention architecture [24]

**Multi-head attention**

A multi-head attention mechanism proposes that self-attention can be run multiple times in parallel mode combining knowledge of the same attention pooling via different representation subspaces of queries, keys, and values. Afterward, the independent attention outputs are concatenated and linearly transformed into the expected dimension, as portrayed by equation 7 and figure 6.

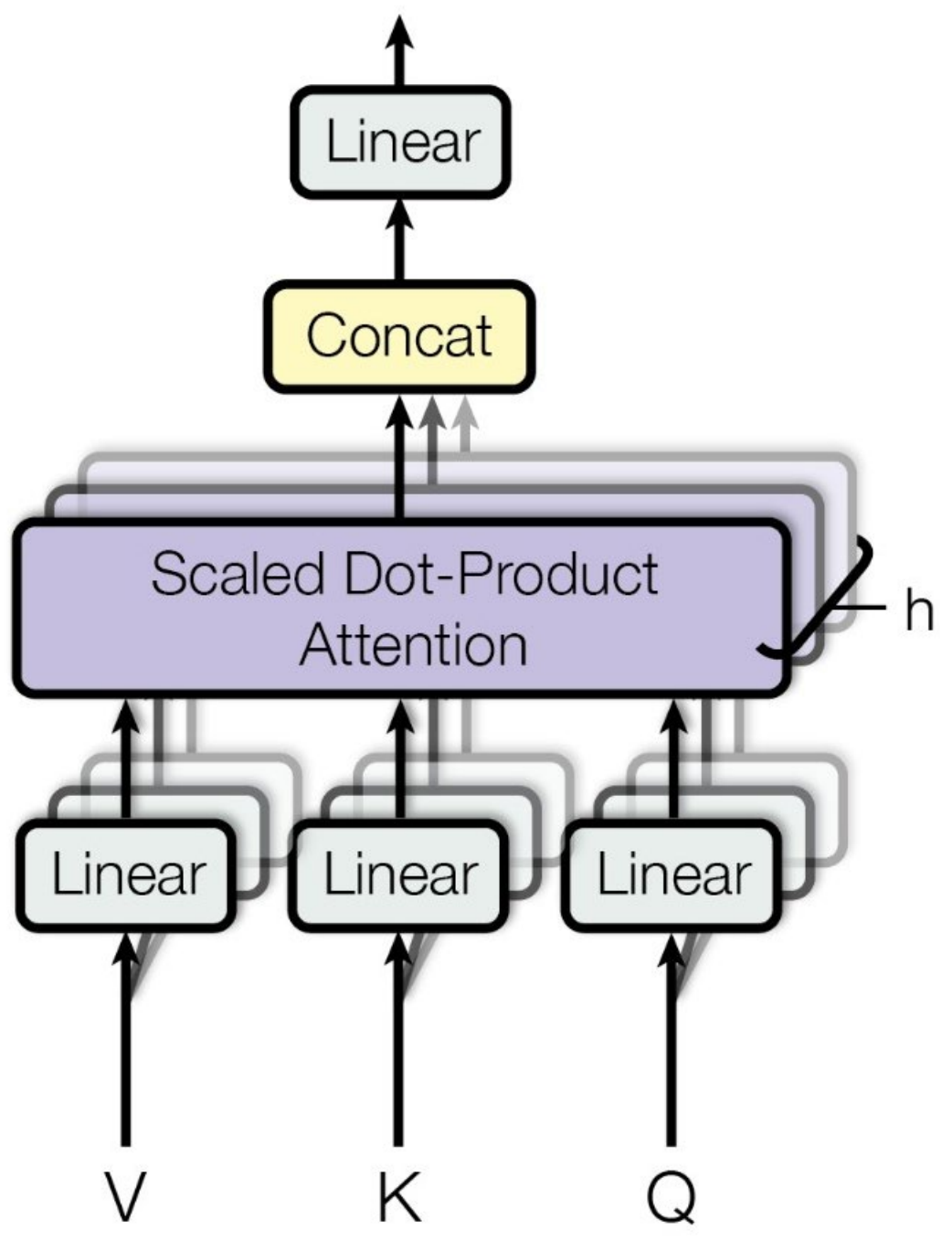

**FIGURE 6.** Self-attention architecture[24]

$$MultiHead(Q,K,V) = Concat(head_1, \ldots head_h)W^o \qquad (7)$$

where $head_i$ = Attention $(QW_i^Q, KW_i^k, VW_i^v)$

Since the Transformer's invention, several variants have been developed to solve different machine-learning tasks in computer vision and natural language processing. It's imperative to note that the state-of-the-art models are built on the foundation transformer architecture [63]. In the following subsection, we will discuss the contemporary generative models.

### 1) GENERATIVE PRE-TRAINED TRANSFORMER (GPT)

A Generative Pretrained Transformer (GPT) describes the transformer-based large language model (LLM) that utilizes deep learning techniques to generate a human-like text [64]. The model was introduced by OpenAI in 2018 [65], following Google's 2017 invention of a transformer. It is made of a stack of transformer decoders. They proposed a model consisting of two stages: learning a high-capacity language model from a large corpus of text and fine-tuning it with labeled data during the discriminative task, as depicted in figure 7.

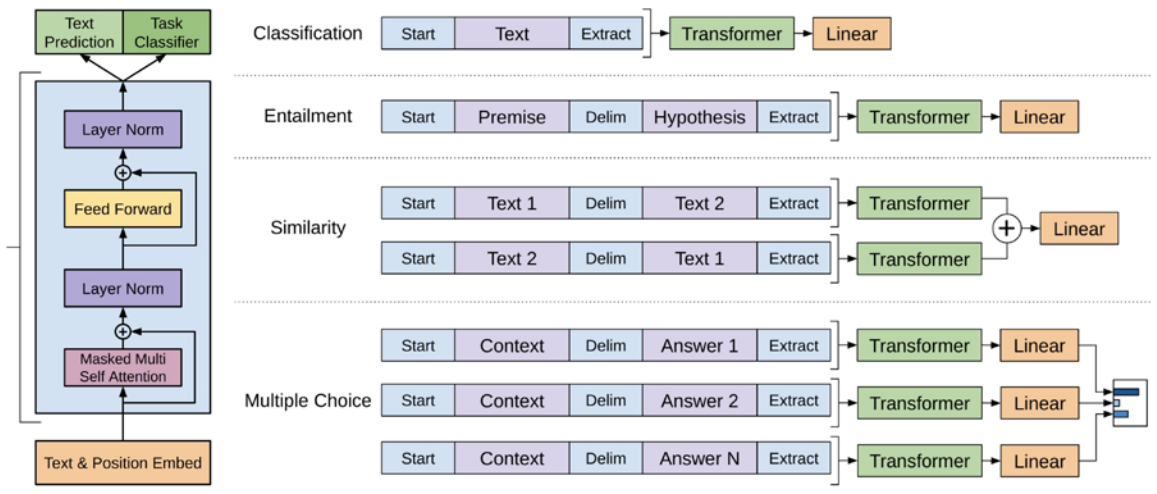

**FIGURE 7.** Self-attention architecture[65]

GPT or GPT-1 was trained on the BooksCorpus dataset, which consists of over 7,000 unique unpublished books in many genres, such as Adventure, Fantasy, and Romance, all with long stretches of contiguous text, allowing the generative model to learn on long-range information [61][62] [65]. The model training specification included the following:

- 12-layer decoder-only transformer.
- Masked self-attention heads (768-dimensional states and 12 attention heads).
- Position-wise feed-forward networks.
- Adam optimization.
- Learning rate: 2.5e-4.
- 3072-dimensional inner states.

The assessment tasks for the model were drawn from four primary categories within Natural Language Processing (NLP): these encompass natural language inference, question answering and common-sense reasoning, semantic similarity, and classification. Following the initial release, OpenAI has produced a series of variant models known as GPT-n series where every successor model is more substantial and efficient than the predecessor. GPT-4 is the most recent variant release in March 2023.

#### 2) GPT-2

After the great success of GPT-1, , OpenAI released a second version (GPT-2) in 2019 with 1.5 billion learnable parameters, ten times more in pre-training corpus and parameters than its predecessor trained on WebText, a collection of millions of webpages. [66]. As a result, this model is capable of handling complex problems and generating coherent and contextually relevant texts across a wide range of topics and styles.

#### 3) GPT-3

This version was released in 2020 and had 2048-token contexts, 175 billion learnable parameters, which is more than 100 times its predecessor, and required 800GB of storage[67]. CommonCrawl was used to train the model, which was tested on all domains of NLP, and it had promising few-short and zero-shot performance. This version was further improved to GPT 3.5, which was used to develop ChatGPT. Considerable research work has been conducted, incorporating GPT-1 to GPT-3.5 across various task such as Speech Recognition [68] [69] [70], Text Generation [71] [72] [73] [74] [75] [76] [77] [78], Cryptography [79] [80] [81] [82], Computer Vision [88] [89], and Question Answering [83] [84] [85] [86] [87].

#### 4) GPT-4

In March 2023, the most recent GPT model was released by OpenAI [90]. It's a multimodal transformer model, A large-scale language model which accepts image and text inputs and produce text outputs. In a number of professional and academic benchmarks, including passing a bar and medical exam at high rates, GPT-4 exhibits high performance comparable to that of humans[91] [92]. The model was trained using publicly available internet data and data licensed from third parties and then fine-tuned using Reinforcement Learning from Human Feedback (RLHF). It was compared with state-of-the-art models using Measuring Massive Multitask Language Understanding (MMLU) [93] that covers 57 tasks in elementary mathematics, US history, computer science, law, and more and outperformed them all.

### C. GENERATIVE ADVERSARIAL NETWORK (GAN)

#### 1) GAN OVERVIEW

A generative adversarial network (GAN) is an unsupervised generative model that consists of two neural networks: a generator and a discriminator. A generator attempts to fabricate new data (fake) that is indistinguishable from real data, while a discriminator tries to distinguish between real and fabricated data [94]. Figure 8 illustrate the schematic architecture of GAN (Also known as a vanilla GAN). The generator network takes noise as input and generates fake data. The discriminator network takes both real and fake data as input and classifies them as real or fake using a sigmoid activation function and binary cross-entropy loss [95]. Since the generator does not have direct access to authentic images ,it only learns through interactions with the discriminator; the discriminator has access to synthetic and authentic images. Upon completion of classification, backpropagation takes place to optimize the training process [94]. This process repeats itself until the difference between real and fake data samples is negligible.

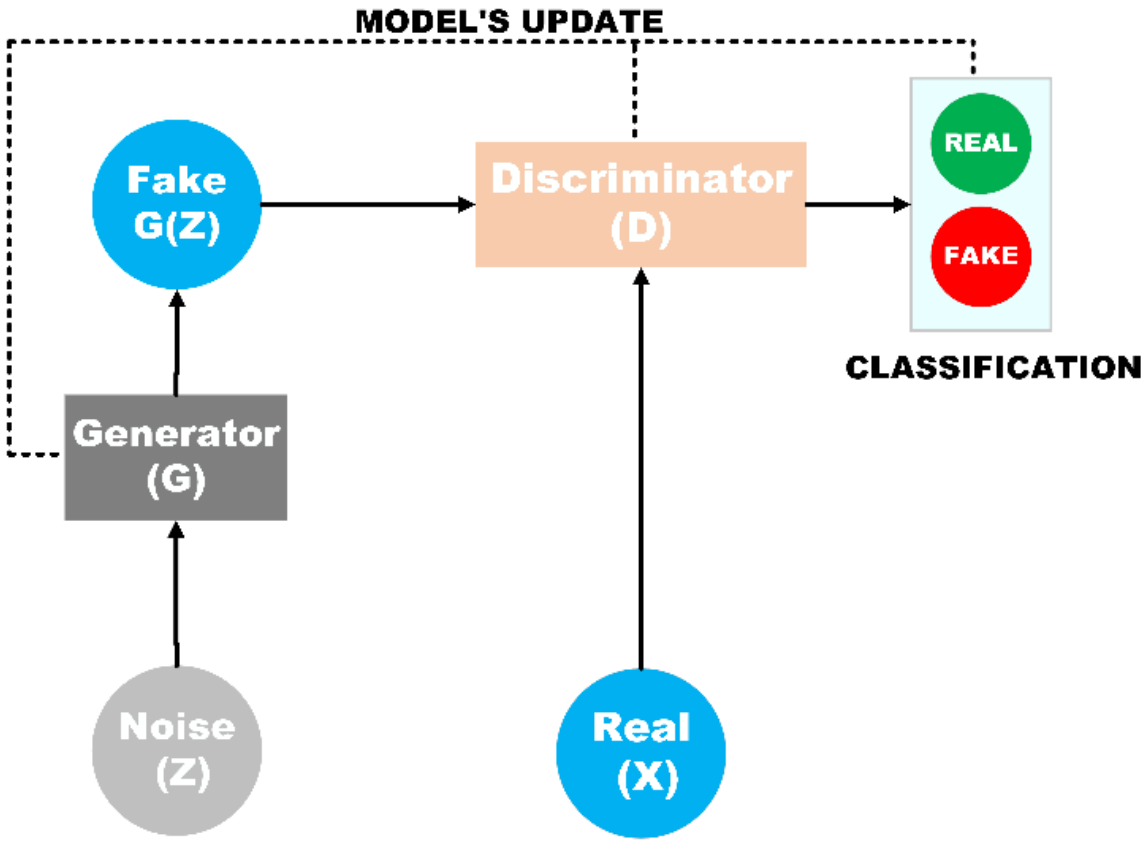

**FIGURE 8.** Schematic GAN architecture

According to Goodfellow et al. [25], the generator (G) and discriminator (D) are trained together in a minimax game

(zero-sum game). In this game as demonstrated by equation 8, G is trying to maximize the probability that D misclassifies its output as real data, while D is trying to minimize the probability that it misclassifies G's output.

$$\boldsymbol{min_G max_D V(D, G)} = \boldsymbol{E_{x \sim p_{data}(x)}[\log D(x)]} + \boldsymbol{E_{z \sim p_z(z)}[\log(1 - D(G(z)))]} \quad (8)$$

Where $\boldsymbol{E}$ is the Expected Value, $\boldsymbol{p_{data(x)}}$ is Real data distribution and $\boldsymbol{p_z(z)}$ implies Noise data distribution.

### 2) GAN CHALLENGES

Despite their robustness, traditional GANs suffer from limitations such as:

**Mode collapse:** In this phenomenon, the generator can only produce a single type of output or a limited number of outputs [96]. This is because the generator becomes stuck in a particular mode or pattern, failing to generate diverse outputs that cover the entire data range [97]. There are two main causes of mode collapse in GANs. The first is catastrophic forgetting [98], which occurs when learning in a current task destroys knowledge learned in a previous task. The second cause is discriminator overfitting, which results in the generator loss vanishing[99].

**Non-convergence and Instability:** The loss function in equation 8 can cause the generator to suffer from gradient vanishing [100]. This can happen when the discriminator learns too quickly and can easily distinguish between real and fake samples. On the other hand, the generator may have a lower learning rate and be unable to keep up. This can lead to the training process stalling, as the generator cannot learn from the feedback provided by the discriminator. GANs are also known to be sensitive to the choice of hyperparameters, such as the learning rate and the batch size. This means that it can be challenging to train GANs consistently, as even small changes to the hyperparameters can significantly impact the results[101].

Gradient vanishing can be addressed using a different loss function, such as the Wasserstein loss. The Wasserstein loss is less sensitive to the discriminator's learning rate, and it can prevent the generator's gradients from disappearing. Another solution would be to use a generator with a smaller learning rate. This will prevent generator weights from becoming too large, which can also contribute to gradient vanishing. In addition, a good initialization technique must be used for the generator. In this manner, the generator will start well, and the training process will likely be successful.

### 3) GAN VARIANTS

In response to the aforementioned GAN challenges, various variants have been developed to address the weaknesses and optimize the model. Here are some of the most famous variants of GAN since its emergence in 2014:

**Conditional Generative Adversarial Network (cGAN)**

cGAN was introduced by Mirza et al. [102] in 2014, this variant enhances the classical GAN by incorporating extra auxiliary information into the Generator and Discriminator networks, such as class labels or style attributes. This integration is achieved by introducing an additional layer that includes the conditional information input to the generator, instructing it on what to produce [103]. For instance, in an image generation scenario, this condition might consist of a class label that precisely defines the type of image to be generated.

**The Deep Convolutional GAN (DCGAN)** framework employs a deep learning model for discriminator and generator components, specifically a Convolutional Neural Network (CNN). In the architectural design defined by Radford et al. [104], traditional fully connected layers situated on top of convolutional features have been omitted. Additionally, including Batch Normalization plays a pivotal role in enhancing training stability. This technique normalizes the input to each neural unit, ensuring a mean of zero and unit variance, thus facilitating more consistent and efficient learning. Moreover, DCGAN substitutes conventional pooling layers with strided convolutions in the discriminator and fractional-strided convolutions in the generator network. The Rectified Linear Unit (ReLU) serves as the activation function for the generator, while the Leaky ReLU is employed in the discriminator. These activation functions play a crucial role in enabling the networks to capture intricate patterns and features.

**Wasserstein GAN** (WGAN) is a GAN variant that employs the Wasserstein distance (also referred to as the Earth Mover's distance) as its loss function, distinguishing itself from traditional GANs that typically use the Jensen-Shannon or Kullback-Leibler divergences. The Wasserstein distance (WD) measures the similarity between the distributions of real and generated samples[105]. It is grounded in the solution to a classical optimization problem known as the transportation problem [106]. In this context, suppose there exists several suppliers, each endowed with a certain quantity of goods, tasked with delivering to several consumers, each having a specified capacity limit. Each supplier-consumer pair incurs a cost for transporting a single unit of goods. The transportation problem aims to identify the most cost-efficient allocation of goods from suppliers to consumers.

$$W(P_r, P_g) = \inf_{\gamma \in \pi(P_r, P_g)} E_{(x.y) \sim \gamma}[\,||\,x - y||\,] \quad (9)$$

WD is expressed by equation 9, $\boldsymbol{P_r}$ *and* $\boldsymbol{P_g}$ denotes the probability distribution of real ad generated sample respectively. The Lipschitz constraint was utilized to impose weight clipping on the discriminator[107]. This measure enhances training stability, mitigating challenges like mode collapse and saturation loss.

**Cycle GAN** is an approach that automates training image-to-image translation models without requiring paired examples, leveraging GAN architecture [108]. It utilizes unassociated image collections from distinct source and target domains (e.g. Domain X and Domain Y). The model structure comprises two generators: Generator-X crafts images for Domain X, and Generator-Y generates images for Domain Y. Each generator associated with a its corresponding discriminator for binary classification.

This variant incorporates three loss functions: firstly, the cycle consistency losses ensure that translations between domains maintain a coherent loop, returning to their original point; secondly, the adversarial loss pits the Generator against its corresponding Discriminator, with the Generator striving to generate domain-specific images while the Discriminator distinguishes between translated and real samples; and thirdly, the Identity Loss incentivizes the Generator to faithfully preserve color composition between input and output, enhancing translation fidelity.

**StarGAN**: a method that harnesses the power of the GAN architecture for versatile multi-domain image-to-image translation. As outlined by Choi et al [109], this innovative generative adversarial network masterfully learns mappings among numerous domains, employing just a single generator and discriminator, and efficiently trains on images spanning all domains. This model utilizes an Adversarial Loss to make generated images virtually indistinguishable from real ones, a Domain Classification Loss to guarantee precise classification by the discriminator and a Reconstruction Loss that minimizes adversarial and classification losses.

In the preceding subsection, we have delved into several variants of Generative Adversarial Networks (GANs). However, it is worth noting that the landscape of GANs encompasses a myriad of additional variants that have significantly advanced beyond the foundational GAN framework. These notable advancements include the Progressive GAN (PGAN) of 2017 [110], BigGAN of 2018 [111], StyleGAN [112] and StyleGAN 2 [113] of 2019, along with earlier innovations such as InfoGAN [114], Stacked GAN [115], Bidirectional GAN (BiGAN) [116] from 2016.

### *D. DIFFUSION MODEL*

Diffusion model is a generative model characterized by a two-step process. Initially, they introduce Gaussian noise into the training data, a step referred to as the forward diffusion process. Subsequently, they perform the reverse diffusion process, often called denoising, to reconstruct the original data. Over time, the model progressively acquires the ability to eliminate the added noise.

## III. GENERATIVE AI TASK

Generative AI encompasses a wide array of tasks, including Speech Generation (Text-to-Speech), Image Generation (Text-to-Image), Text Generation (Text-to-Text), Code Generation (Text-to-Code), Music Generation (Text-to-Music), Video Generation (Text-to-Video), and Scientific Content Generation (Text-to-Science). These tasks are supported by various cutting-edge tools, as illustrated in Table 2. Notably, Google boasts the most generative tools, with Meta AI and OpenAI following closely. Most of these tools were unveiled in 2023, although a few were introduced earlier.

### *A. TEXT GENERATION*

This task involves taking text as input and generating corresponding text-based responses. It is often associated with question-and-answer conversational systems, commonly called chatbots. Many renowned generative AI tools fall within this category, with ChatGPT being a groundbreaking example in the field of Generative AI. Other notable tools in this category include Google's Bard, OpenAI's ChatGPT Plus, Wordtune Spice, and Cohere's Generate. We conducted a comprehensive performance assessment of two prominent and renowned text-to-text tools, Bard and ChatGPT. Both were presented with identical queries: '*Provide a brief description of what Bard is in one paragraph*', '*Provide a brief description of what ChatGPT is in one paragraph*', and a Swahili question, '*Habari za saa hizi*'. The results as illustrated by figure 9, unmistakably indicate that ChatGPT outperformed Bard in delivering more precise answers to the questions.

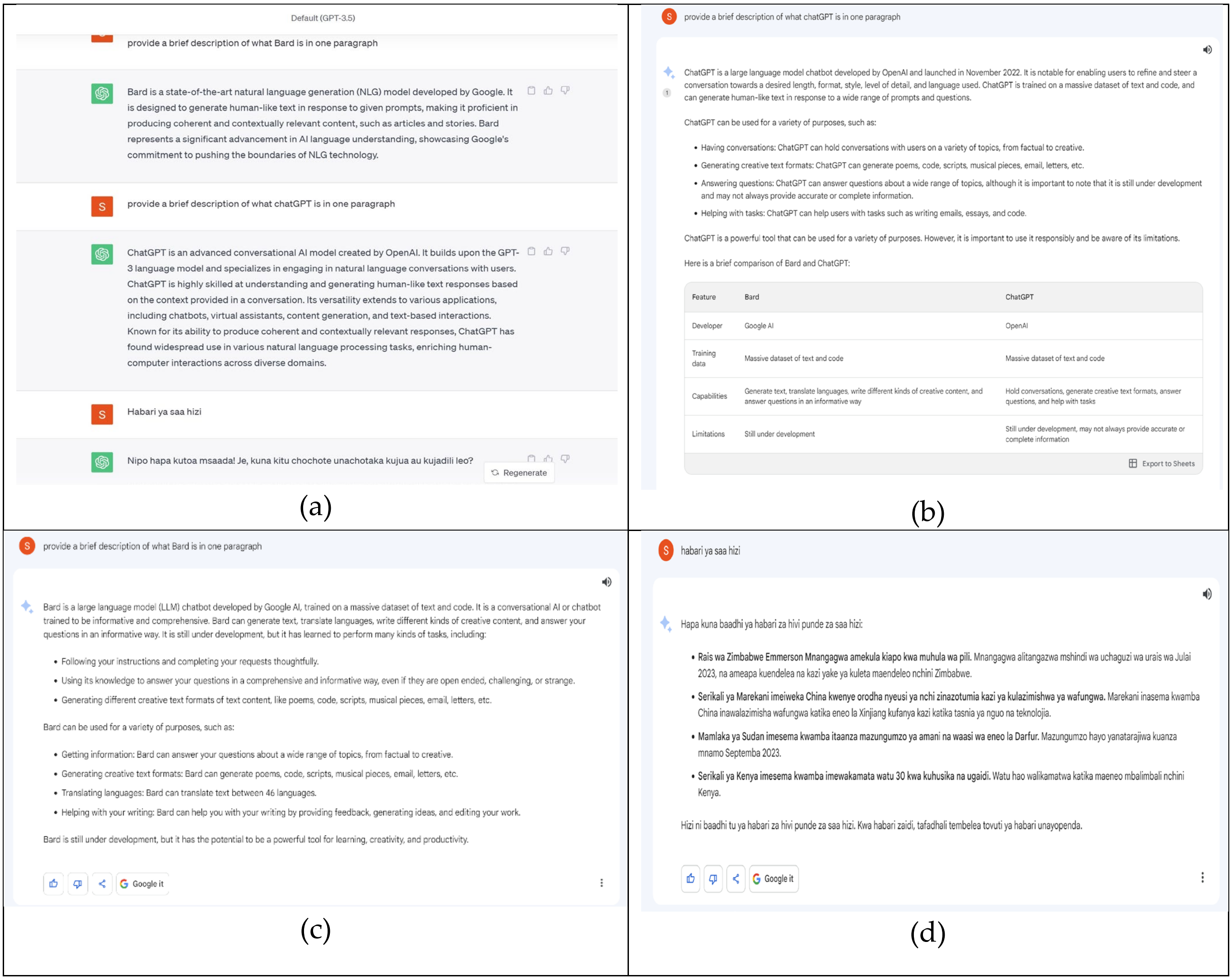

**FIGURE 9.** (a) chatGPT chatbot, b-d are bard output.

### B. IMAGE GENERATION

It's a task which encompasses the process of utilizing textual prompts or visual to generate corresponding images, spanning various visual domains, including graphics, photographs, and artwork. As an illustration of text-to-image concept, we conducted experiments using 'Firefly' from Adobe and 'Stable Diffusion' by Stability as our subjects. By prompting these models with '*College Student Programming*', we obtained their respective outputs, as showcased in Figure 10, the results clearly indicate that while 'Firefly' excelled in delivering more precise outputs in alignment with the input, Stable Diffusion exhibited superior image resolution compared to its counterpart. Another scenario image generation revolves around the transformation of an image from one form to another, guided by textual descriptions provided as input. Within this domain, numerous tools have demonstrated promising capabilities in effecting such transformations. Notably, we have explored the performance of RoomGPT and Runaway, as exemplified in Figure 11 and Figure 12, respectively.

### C. VIDEO GENERATION

This task involves generating new videos based on textual or visual inputs, whereby visual encompasses a diverse range of content that includes both images and videos. In this domain, there are notable tools designed to accommodate exclusively text-based descriptions as inputs. A prime example is 'Parti' by Google, and DALL E-2 [117] by openAI are proficient tools focused on creating videos solely from textual prompts. Nonetheless, the field of video generation is in a state of continuous evolution. Tools such as 'Gen-2' by RunwayML, 'Imagen Video' by Google[118], and 'Make-A-Video' by Meta[119] have emerged as pioneers. These advanced platforms possess the remarkable capability unlimited to textual descriptions but also seamlessly integrate images and videos as input, transcending conventional boundaries. Their excellence lies in their adeptness at transforming these inputs into entirely novel video compositions, thus unveiling the exciting potential of generative AI in the creative realm of video production.

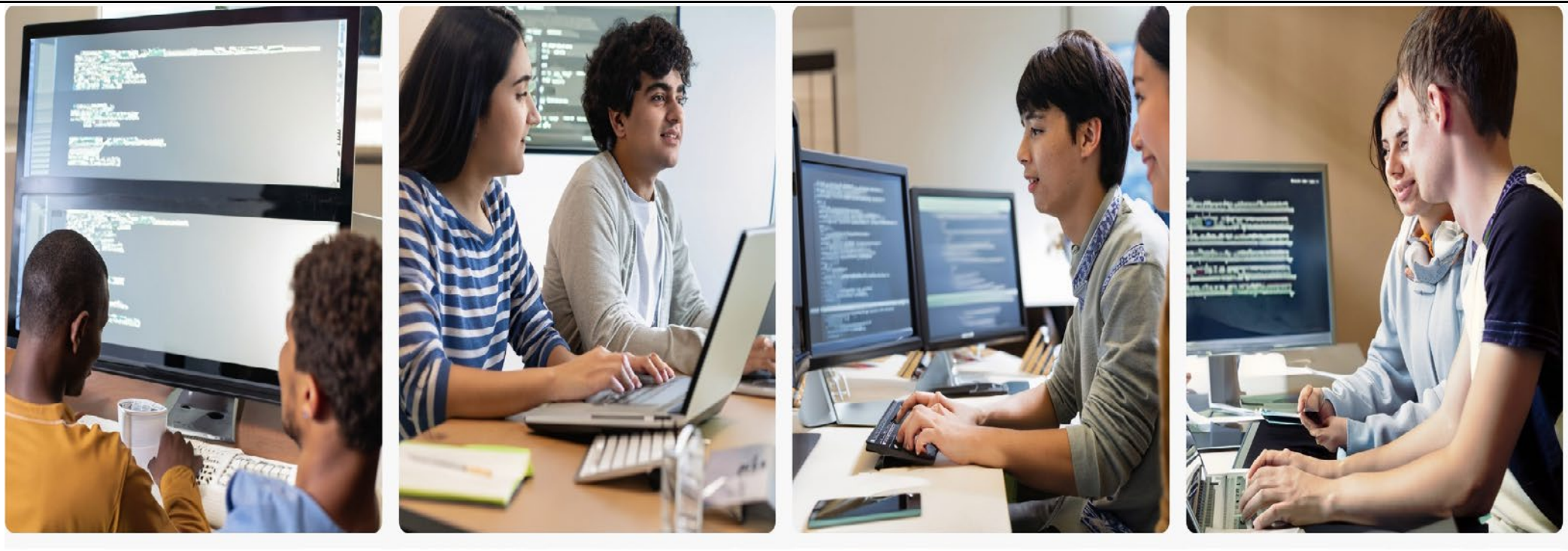

(a)

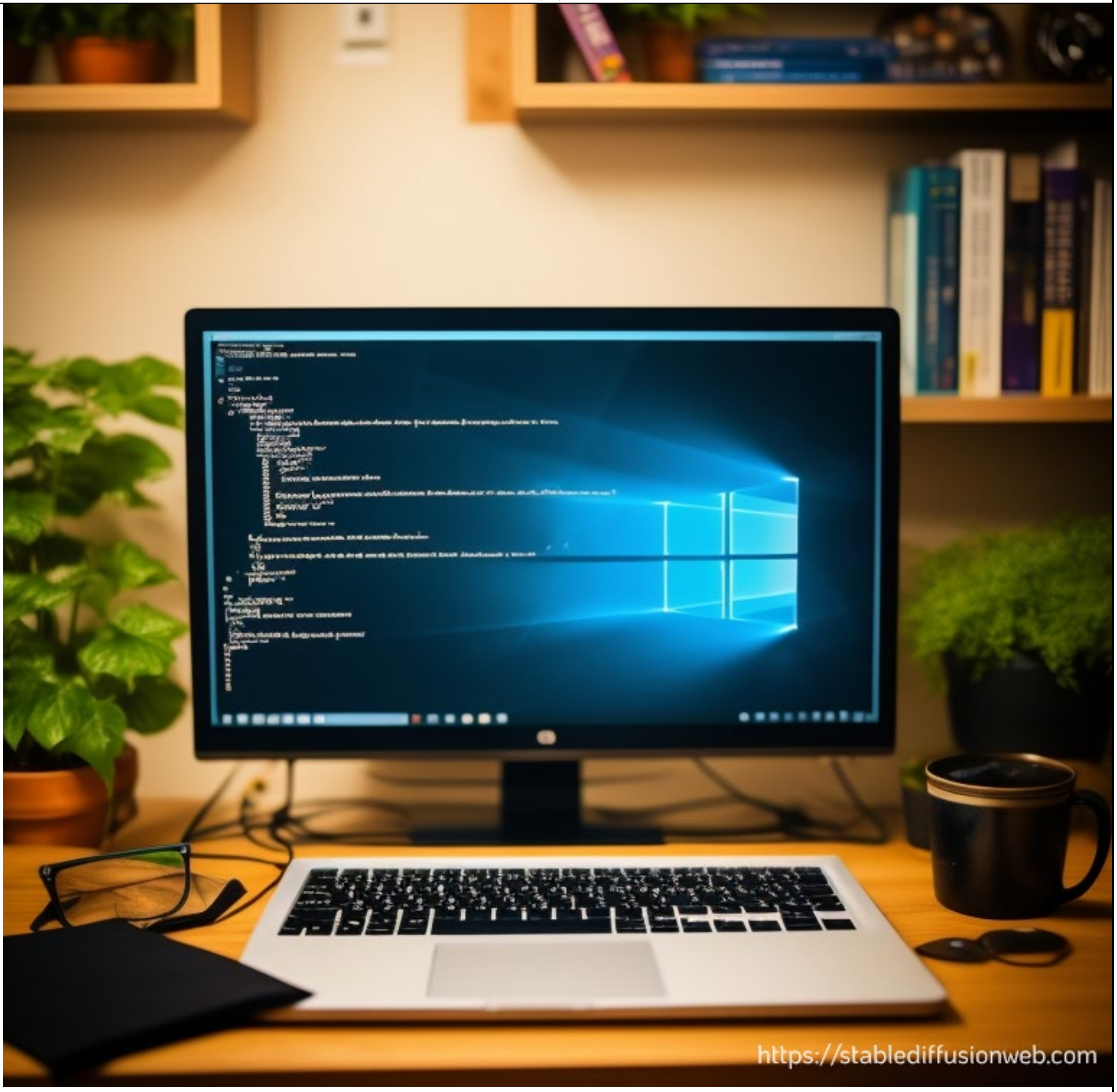

(b)

**FIGURE 10.** (a) Adobe firefly, (b) stable diffusion image generated using text "*college Student Programming*".

### D. CODE GENERATION

Code generation tools are specialized software utilities capable of automatically producing code blocks for various programming languages based on textual descriptions provided as input[120]. These tools leverage sophisticated models trained on extensive publicly available code repositories, boasting billions of parameters. Their primary objective is to assist human developers by comprehending plain English and translating it into functional code. Notable examples of such tools include StarCoder [121], Codex [122], CoPilot, Codey, and Code Interpreter. Additionally, it's worth noting that several text-to-text tools, including ChatGPT and Bard as depicted by Figure 13 , also possess the capacity to generate code.

### E. MUSIC GENERATION

It's a fascinating generative task involving entirely new music's composition. This innovative process takes input in various forms, including textual descriptions, sequences of musical notes, and even audio samples[123]. The objective is to harness these inputs and transform them into fresh musical compositions that encapsulate rhythm, melody, harmonious chords, and diverse musical instruments. Prominent tools like MuseNet [124] and Jukebox[125] stand out as prime examples in the music generation. These innovative platforms harness the power of generative AI to craft musical compositions spanning various genres and styles. They excel in infusing creativity into the art of music, opening new avenues for artists and enthusiasts to explore and enjoy.

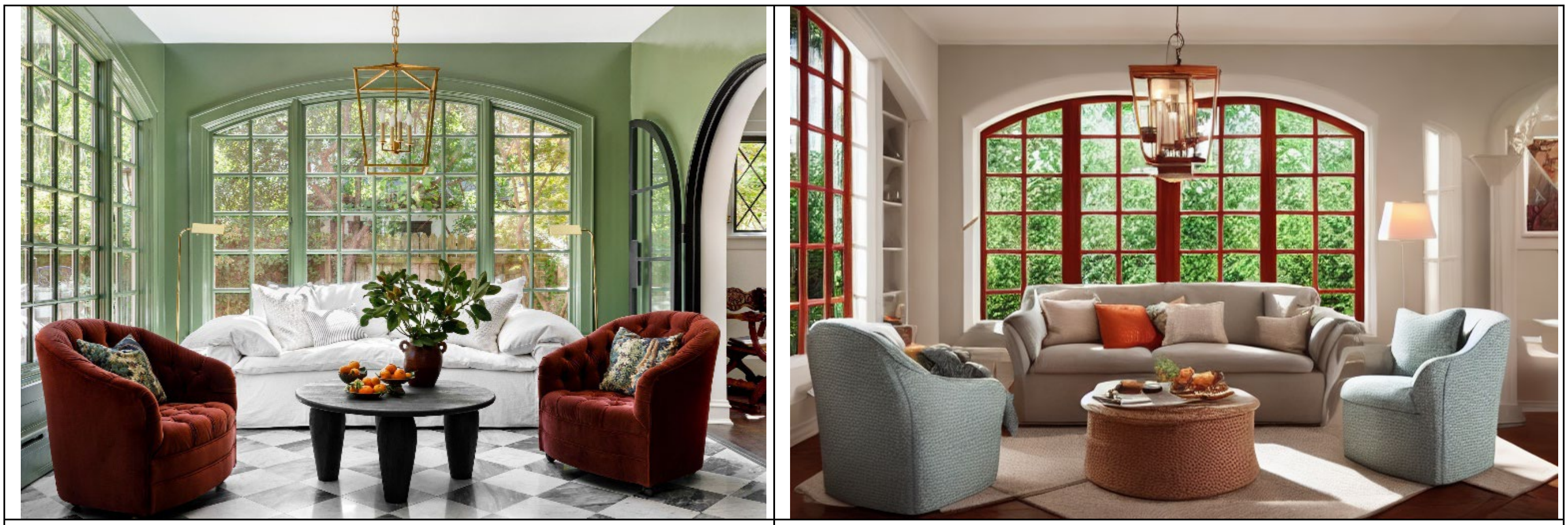

a b

**FIGURE 11.** (a) Original Living Room [126] and (b) Generated new living room using roomGPT

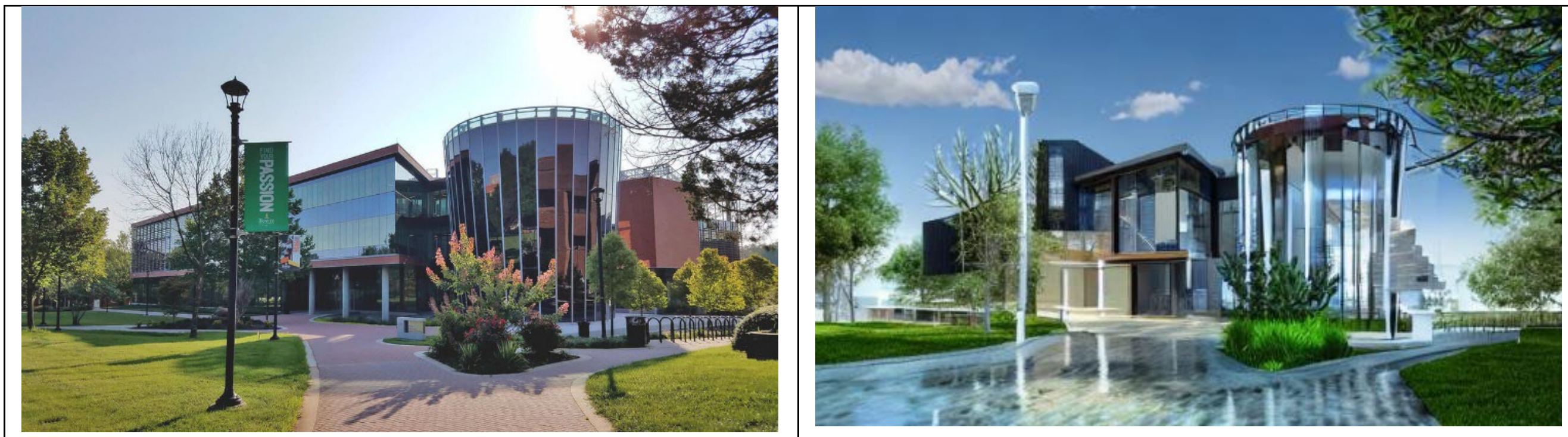

a b

**FIGURE 12.**(a) Original BSU Natural Science Building[127] and (b) Generated new living room using runway with prompt

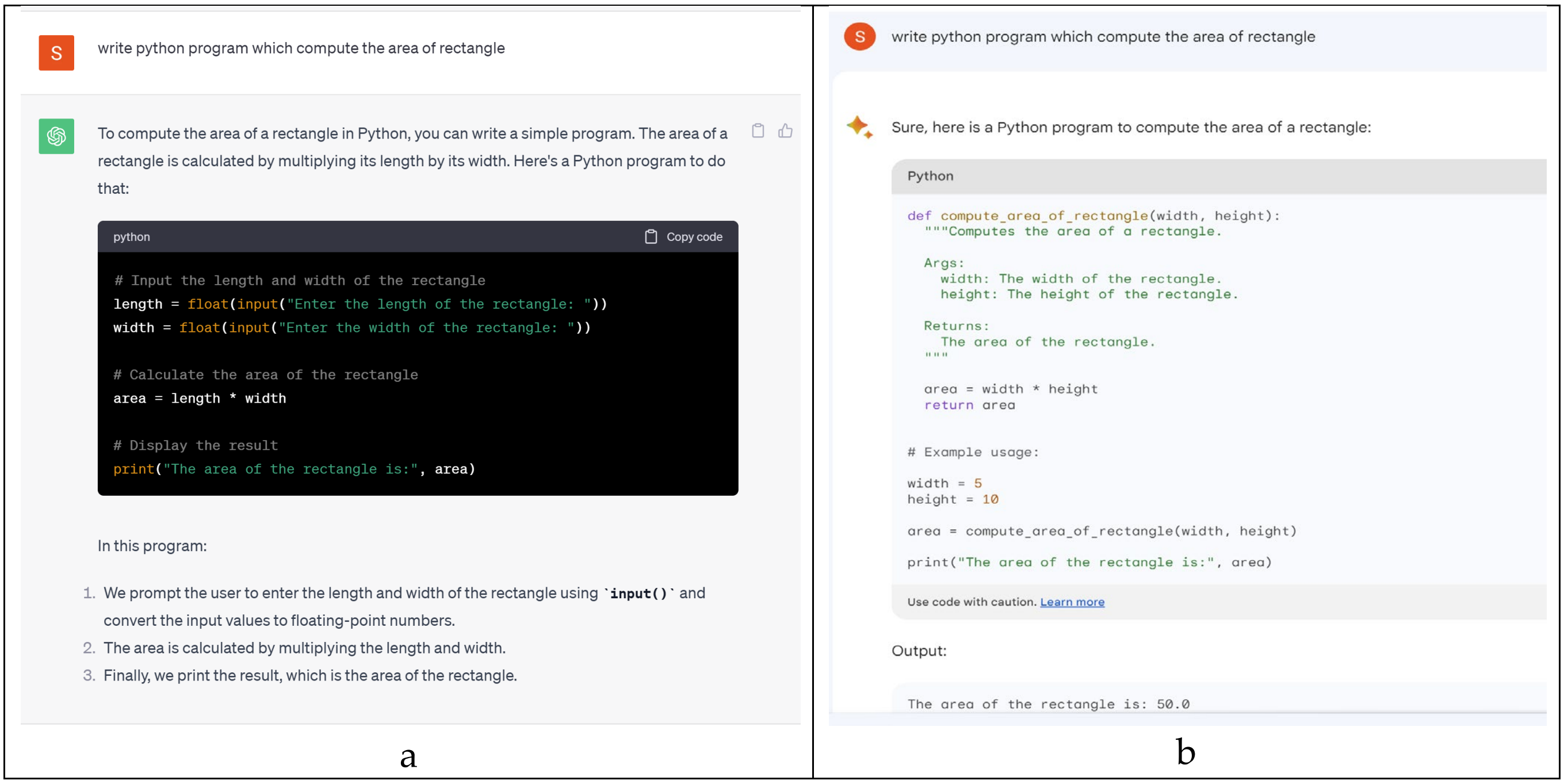

**FIGURE 13.** Code generation using (a) ChatGPT and (b) Bard

### F. SPEECH GENERATION

The generation of human-like speech or voice relies on textual or audio input. Textual input can encompass written text, such as sentences, paragraphs, or entire documents, and it can span multiple languages, including punctuation, special symbols, and formatting instructions. Speech generation models, such as SpeechGAN, undertake a sequence of steps that involve speech synthesis, enhancement, and conversion. The enhancement process includes noise handling, tone modulation, emotion conveyance, and other nuanced features[128][129]. Numerous tools have been developed in this domain to facilitate speech generation, some of which include Whisper, Speechelo, Synthesys, Voice Over, and WaveNet. These tools are proficient in generating voices or speech that closely mimic natural language, effectively blurring the line between human and artificial speech synthesis.

### G. SCIENTIFIC CONTENT GENERATION

Scientific content generation is a multifaceted process encompassing the creation of informative and scholarly content across various domains of science, including mathematics, physics, chemistry, and biology. This endeavor seeks to harness the power of generative AI to produce content that is accurate and insightful, aiding in disseminating scientific knowledge. One notable study in this field, conducted by Rodriguez et el.[130], delved into the innovative way of generating scientific figures based on textual input. This groundbreaking research leveraged diffusion models to seamlessly translate textual descriptions into visually informative scientific figures, thereby streamlining the process of scientific communication and visualization. Furthermore, Google's ongoing research project, Minerva[131], represents a significant stride in solving quantitative reasoning problems. This initiative harnesses the capabilities of Large Language Models (LLMs) to tackle complex quantitative challenges, thereby enhancing our understanding of mathematics and its practical applications within the scientific landscape. In parallel, Galactica [132], a cutting-edge tool developed by Meta AI, plays a pivotal role in scientific writing. This platform equips scientists and researchers with powerful tools to streamline articulating their scientific discoveries, theories, and insights.

TABLE 2
GENERETIVE AI TOOLS

| | Tool | Developer | Task | Year | Additional Description |
|---|---|---|---|---|---|
| 1 | VoiceBox | Meta AI | Text-to-Speech | 2023 | Generate voice clips |
| 2 | Genny | Lovo | Text-to-Speech, Text-to-Image | 2020 | Can generate voice over and art image |
| 3 | Metamate | Meta AI | Text-to-Code | 2023 | Software debugging |
| 4 | Scribe AI | Scribe | Computer Vision | 2023 | Creates Documentation, how-to guides, SOPs and training manuals |
| 5 | Read | Read.ai | Speech-to-Text | 2021 | Virtual meeting Automated summary, transcripts, playback, and highlights on action items, key questions, and real-time engagement |
| 6 | appleGPT | Apple | Text-to-Text | 2023 | chatbot summarize text and answer questions |
| 7 | Einstein GPT | SalesForce | Text-to-Text | 2023 | Chatbot built in top for chatGPT which generate text, translate languages, write different kinds of creative content, and answer questions |
| 8 | flashGPT | Neuroflash | Text-to-Text | 2020 | A generative Chatbot which use flash |
| 9 | AlphaCode | DeepMind | Text-to-Code | 2022 | Generate code,creative content, and respond to questions in an informative way |
| 10 | Cloude 2 | Anthropic | Text-to-Text | 2023 | Content Generation, AI Assistant |
| 11 | Jasper | Jasper | Text-to-Text | 2021 | Generate Creative Contents |
| 12 | PaLM 2 | Google | Text-to-Text | 2023 | Generate code, creative content, Translation and Q&A |
| 13 | Shepherd | Meta AI | Text-to-Text | 2023 | Improve the accuracy of AI generated response |
| 14 | Murf | Murf.Ai | Text-to-Speech | 2020 | Generate voice-over for Creative contents and Presentation |
| 15 | Codex | OpenAI | Text-to-Code | 2021 | Code Generator |
| 16 | Codey | Google | Text-to-Code | 2023 | Generate Code based on user input |
| 17 | DALL-E 2 | OpenAI | Text-to-Image | 2023 | Generate image from text description |
| 18 | DeepDream | Google | Text-to-Image | 2015 | Generate psychedelic images |
| 19 | Midjourney | Midjourney, Inc | Text-to-Image | 2022 | Generate realistic and creative image from text prompt |
| 20 | Firefly | Adobe | Text-to-Image | 2023 | Generative image from text prompt |
| 21 | RoomGPT | RoomGPT.io | Text-to-Image | 2023 | Design home and room |
| 22 | StyleGAN | Nvidia | Text-to-Image | 2019 | Generate realistic and creative image from text prompt |
| 23 | Stable diffusion | Stability AI | Text-to-Image | 2022 | Generate photo-realistic images given any text input |
| 24 | NovelAI | Anlatan | Text-to-Image | 2021 | Generate image from text input and strorywriting |
| 25 | CM3leon | Meta AI | Text-to-Image | 2023 | generate text and images |
| 26 | Imagen | Google | Text-to-Image | 2023 | Generate realistic image |
| 27 | Photosonic | Writesonic | Text-to-Image | 2020 | Generate image from text input |
| 28 | AI Art | Nightcafe | Text-to-Image | 2019 | Generate image from text input |
| 29 | Canva AI | Canva | Text-to-Image | 2023 | Generate image from text input |
| 30 | Dreamstudio | Stability AI | Text-to-Image | 2022 | Generate photo-realistic images given any text input |
| 31 | StarryAI | StarryAI Inc | Text-to-Image | 2021 | Generate image from text input |
| 32 | ChatSonic | Writesonic | Text-to-Image, Text-to-Text, | 2022 | Conversational chatbot which can generate human text response and image |
| 33 | Soundful | soundful | Text-to-Music | 2021 | Create customized music based on individual needs |
| 34 | Boomy | Boomy | Text-to-Music | 2019 | Develop music without prior knowledge. |
| 35 | Soundraw | Soundraw Inc | Text-to-Music | 2021 | Generate Music |
| 36 | AudioCraft | Meta AI | Text-to-Music | 2023 | Music Generator |
| 37 | MusicGen | Meta AI | Text-to-Music | 2023 | Music Generator |
| 38 | Galactica | Meta AI | Text-to-Science | 2022 | tool for scientific writing |
| 39 | Minerva | Google | Text-to-Science | 2022 | Solve Quantitative reasoning problem |
| 40 | WaveNet | DeepMind | Text-to-Speech | 2016 | Generate realistic speech from text or other audio inputs |
| 41 | Voice Over | Speechify | Text-to-Speech | N/A | Creates natural Voiceovers for any Content |
| 42 | TexTalky | Textalky | Text-to-Speech | 2021 | Creates realistic voice from text |
| 43 | speechelo | speechelo | Text-to-Speech | | Creates realistic voice from text |
| 44 | Overdub | Descript's | Text-to-Speech | 2021 | Creates realistic voice from text |
| 45 | Synthesys | Synthesys | Text-to-Speech | 2020 | Create voiceover from text |
| 46 | Kits | kits | Text-to-Speech | N/A | Voice generator |
| 47 | WellSaid | WellSaid Lab | Text-to-Speech | N/A | Voice generator |
| 48 | Altered Studio | Altered | Text-to-Speech | 2023 | Voice generator |
| 49 | Whisper | OpenAI | Text-to-Speech | 2022 | Speech recorgnition and translation |
| 50 | Jukebox | OpenAI | Text-to-Speech | 2020 | Music Generator |
| 51 | LaMDA 2 | Google | Text-to-Speech | 2022 | Customer Service Chatbots, Q&A, Translation, Research |
| 52 | PEER | Meta AI | Text-to-Speech | 2022 | Writing tool |
| 53 | chatGPT | OpenAI | Text-to-Text | 2022 | Conversational chatbot that generates human-like text responses |
| 54 | Bard | Google | Text-to-Text | 2023 | Conversational chatbot that generates human-like text responses |
| 55 | Generate | Cohere | Text-to-Text | 2022 | Content Generation |
| 56 | chatGPT plus (GPT-4) | OpenAI | Text-to-Text | 2023 | Advanced ChatGPT, Conversational chatbot that generates human-like text responses |
| 57 | Wordtune Spice | AI21 Labs | Text-to-Text | 2023 | Writing Generator |
| 58 | Gen-2 | RunwayML | Text-to-Video | 2023 | Design video from text input |
| 59 | Synthesia | Synthesia | Text-to-Video | 2018 | Generate video from text input |
| 60 | Make-A-Video | Meta AI | Text-to-Video | 2022 | Generate video from text input |
| 61 | Imagen Video | Google | Text-to-Video | 2022 | 1280x768 HD videos at 24 frames per second from text limited to inanimate objects |
| 62 | Phenaki | Google | Text-to-Video | 2023 | Generate video from text input of animate objects |
| 63 | Descript | Descript | Text-to-Video | 2020 | Generate video from text input |
| 64 | GitHub Copilot | Microsoft/GitHub/OpenAI | Text-to-Code | 2021 | Code Generator and Suggestion |
| 65 | Sensei | Adobe | Text-to-Image | 2017 | Generate automative workflow and personalize cunstomer experience |
| 66 | parti | Google | Text-to-Image | 2023 | |
| 67 | StarCoder | Hugginface + ServiceNow | Text-to-Code | 2023 | state-of-the-art large language model (LLM) for code |
| 68 | Amper | Amper | Text-to-Music | 2023 | Generate Music of various genres |
| 69 | MuseNet | OpenAI | Text-to-Music | 2019 | Generate Music of various genres |
| 70 | MusicLM | Google | Text-to-Music | 2023 | Generate Music of various genres |
| 71 | quillbot | Course Hero | Text-to-Text | | Can paraphrase, rewrite the text |
| 72 | Rephrase.ai | Rephrase.ai | Text-to-Video | | Can Generate video using avatar by text prompt |
| 73 | Studio bot | Google | Text-to-Code | 2023 | Code Companion for android developer |

## IV. INDUSTIRIAL APPLICATION OF GENERATIVE AI

Generative AI technology's relevance in the present and future is indispensable. Currently, Generative AI is exerting an exponential impact across a broad spectrum of industries, and this section will delve into a detailed exploration of the sectors that are mostly impacted.

### A. MEDIA AND ENTERTAINMENT

In the entertainment industry, Generative AI models are beginning to have a significant impact despite being in their early stages. Their influence spans various entertainment domains, encompassing scriptwriting and storyboarding for novels, plays, and films, audio production[133] involving composition, arrangement, and mixing, game design and character creation, the creation of captivating virtual worlds, marketing campaigns, and the generation of both moving and static images. Notably, a wide range of accessible tools, as demonstrated in Table 3, make it easier to generate content such as reels, jokes, and images[134]. Many of these tools are cost-effective or even free, providing an alternative to traditional content creation methods. As an illustration of their potential, in 2022, RunwayAI played a role in creating the Academy Award-winning film "Everything Everywhere All at Once" which received recognition with seven Oscars award[135] [136].

### B. EDUCATION AND RESEARCH

Generative AI is rapidly reshaping the educational landscape, offering innovative solutions that elevate the learning experience for both students and educators. One significant impact of Generative AI in education is the emergence of personalized content generation tools. Exemplified by technologies like GPT-3, GPT-4 and Bard, these tools empower educators to craft tailored learning materials, including interactive lessons, quizzes, and study guides, precisely catering to the unique needs of individual students and instructors[137]. Furthermore, AI-driven chatbots and virtual tutors provide students with real-time support, offering explanations, addressing queries, and delivering personalized feedback[138]. This transformative technology holds the potential to reinvent how students access and engage with educational content, promoting accessibility and adaptability according to each learner's specific preferences[139] [140].

Generative AI has also opened new avenues of research and academic exploration. The rapid development of Generative AI tools has piqued the interest of researchers and academics across the globe, leading to an array of research opportunities [141]. Tech giants and research institutions are investing significant resources to explore and invent new tools and technologies in this field. This is evident in the surge of publications related to Generative AI, both in peer-reviewed databases like IEEE and non-reviewed platforms like arXiv, where Generative AI topics have gained prominence. The fusion of education and Generative AI has not only transformed the learning experience but has also sparked a thriving academic domain that promises continued growth and innovation[142].

### C. HEALTHCARE

Generative AI is making substantial inroads in healthcare, particularly in medical imaging[143]. It plays a crucial role in overcoming challenges related to limited datasets by enabling the synthesis of new data[144] [145], ultimately enhancing the quality and diversity of medical images. This innovation is set to revolutionize disease detection and diagnosis, providing healthcare professionals with more accurate and detailed information. In addition, Generative AI is transforming the administrative aspects of patient care. By streamlining administrative processes and offering virtual health assistants, it simplifies healthcare management and provides personalized health advice, medication reminders, and emotional support[146].Moreover, Generative AI is revolutionizing treatment planning. Leveraging patient-specific data, it can generate customized treatment plans tailored to an individual's genetic makeup, lifestyle, and medical history. This approach represents a significant leap toward precision medicine, ensuring patients receive the most effective and personalized treatment.

Furthermore, Generative AI is playing a pivotal role in the realm of drug development and discovery[147] [148] [149]. Through the generation of molecular structures[150] and predictive modeling, it expedites the identification of novel therapeutic compounds. These advancements can address previously untreatable diseases, instilling hope in countless patients across the globe. Notably, the collaboration between NVIDIA and Evozyne in implementing Generative AI, specifically ProT-VAE, signifies the remarkable synergy between AI and the healthcare sector. By employing the Protein Transformer Variational AutoEncoder, they have laid the groundwork for creating synthetic proteins[151], opening up new avenues for therapeutic solutions in the fight against challenging incurable diseases. Yet another noteworthy example is the collaborative research venture between Google and Cognizant[152]. Their joint effort aims to construct a Large Language Model (LLM) tailored for healthcare applications, specifically focusing on enhancing Healthcare administrative tasks. This endeavor harnesses the capabilities of Google Cloud and its framework to create cutting-edge generative AI solutions for the healthcare sector.

### *D. BUSINESS*

Generative AI has firmly established its presence in the business landscape. Many of the applications listed in Table 3 operate on a subscription-based model, reflecting the growing commercial nature of these tools. Bloomberg Intelligence predicts that Generative AI (GAI) will generate $137 billion in 2023 and is expected to surge to $1.3 trillion by 2030[153]. This profound impact extends across various industries, from manufacturing and wholesale to retail businesses, banking, agriculture, and many more. Generative AI's reach spans from creating new products and automating financial data analysis to generating personalized advertising campaigns[154][155] [156], offering tailored product recommendations to customers, and producing product descriptions and news articles[157]. It is increasingly evident that Generative AI is reshaping the business landscape and holds immense economic potential in the future.

For example, Amazon is actively harnessing Generative AI capabilities to empower sellers in crafting engaging, compelling, and effective product listings through brief descriptions of their products. Amazon leverages Generative AI to generate high-quality content, which sellers can further refine or directly submit to enrich the Amazon catalog[158].

## V. THE FUTURE OF GENERATIVE AI

Generative AI undoubtedly holds a significant and promising future, offering a plethora of tangible and transformative possibilities across various domains. However, it is equally accompanied by a considerable degree of uncertainty and a range of concerns that deserve in-depth exploration. This section aims to explore the multifaceted aspects of Generative AI, addressing its potential as well as the challenges and uncertainties that lie ahead.

### *A. PIONEER OF FIRFTH INDUSTRIAL REVOLUTION (5IR)*

Generative AI represents the promising frontier of the fifth industrial revolution (5IR), a force poised to revolutionize the fourth industrial revolution and create transformative changes across various sectors. This transformation is made possible by the profound interconnection of internet infrastructure, extensive datasets, and distributed computing resources that transcend geographical boundaries. Several industries, including Healthcare, Security, Cyber Infrastructure, Entertainment, and Education, are on the verge of significant disruption due to Generative AI's capabilities. However, it's crucial to recognize that this disruptive potential may also bring about infrastructure reforms across multiple sectors, potentially leading to high levels of automation and optimization in various career fields.

**On Healthcare Industry**, as we have witnessed, Generative AI is already playing a pivotal role in drug discovery, with a particular emphasis on exploring protein molecules. The potential for this technology in the field of drug development is vast, and substantial investments from major technology companies underscore the anticipated advancements in the near future. However, the impact of Generative AI extends far beyond drug development, as it is expected to transform the patient experience within the healthcare sector fundamentally. By harnessing patients' medical history data, it can autonomously diagnose medical conditions by analyzing metadata like age, sex, and underlying medical conditions. Moreover, it can sift through extensive patient data to identify patterns, make predictions, and suggest appropriate medications. This transformation is set to prioritize patient-centered clinical experiences and drive cost-effectiveness, ultimately leading to significant enhancements in healthcare protocols*[159]*.

**Enhanced Entertainment,** In the foreseeable future, we stand at the threshold of a transformative era where generative AI will likely dominate the realm of content creation in entertainment and media. From crafting intricate scripts and narratives to meticulously arranging scenes and bringing characters to life, the influence of generative AI is set to permeate every facet of content generation in these industries. Furthermore, the potential impact is so profound that it might even challenge the boundaries of life and art. Deceased artists could potentially continue to release new albums and creative works, effectively transcending the limitations of mortality. Not only will this innovation usher in a new age of artistic exploration, but it also promises significant cost savings, revolutionizing the economics of movie and music production. Automating scene creation and content generation will reduce expenses and make the creation process more efficient.

**New education era,** the advent of AI chatbots like ChatGPT and Google Bard, along with other innovative tools, serves as compelling evidence of the democratization of Generative AI in the education industry. This remarkable progress has rendered the current educational system and resources outdated, particularly in developed countries. It anticipates a comprehensive overhaul of the education system, including teaching resources, to adapt to the exponential growth in the generative AI era, aiming to provide highly personalized and adaptive learning experiences.

**Advanced Manufacturing Industries,** before the emergence of Generative AI, robotics had already showcased impressive capabilities. However, with the integration of generative AI, we can look forward to truly remarkable advancements. Just envision the consequences of infusing generative AI into military technology, where we might see the development of generative nuclear weaponry, the formulation of chemical recipes for beverages, detergents, and various industrial products, and the widespread adoption of self-driving vehicles. The range of possibilities is extensive, and it undoubtedly signifies the onset of a new era—an industrial revolution that promises a thoroughly transformed landscape and innovative approaches across numerous sectors of industries.

### *B. JOB MARKET SHIFTING*

The influence of Generative AI on the labor market is two-fold:

Firstly, it ushers in **new employment opportunitie**s in emerging domains such as AI Explainability and Generative AI engineering. McKinsey's analysis ***[160]*** suggests a gradual rise in job openings within professions exposed to Generative AI, and this trend is expected to persist until roughly 2030. A noteworthy revelation is that a substantial 84% of the U.S. workforce occupies positions with the potential to leverage Generative AI for automating a significant portion of repetitive tasks, leading to a considerable surge in overall productivity. Significantly, 47% of U.S. executives express confidence that integrating Generative AI will lead to heightened productivity across diverse industries***[161] [162]***.

Conversely, **Job deterioration**; optimizing and automating business processes are anticipated to replace many existing careers with creative and generative AI functions. Generative AI's impact on the labor market is poised to transform the employment landscape, gradually replacing many traditional roles with advanced technology. According to the World Economic Forum's report***[163]***, tasks with the highest potential for automation by Large Language Models (LLMs) are routine and repetitive. These tasks include those performed by Credit Authorizers, Checkers, Clerks, Management Analysts, Telemarketers, Statistical Assistants, and Tellers***[164] [165]***. Therefore, individuals must prioritize reskilling and adaptability to prepare for AI-driven jobs in the future effectively.

### *C. PRIVACY AND SECURITY CONCERNS*

The cybersecurity infrastructure domain is presently undergoing a profound and rapid transformation, primarily driven by the integration of Generative AI. This substantial shift is giving rise to a host of pressing concerns and challenges for the future:

**Sophisticated cyberwarfare**, currently, we are witnessing a notable surge in malicious activities, and this trend is expected to continue its upward trajectory while also becoming more intricate and sophisticated[166]. For instance the emergence of cutting-edge cyber threat tools like WormGPT and FraudGPT[167] [168], which have rapidly established themselves as pioneering elements in cyber threats often referred to as “exclusive bots” [169] by their perpetrators, are engineered to be highly sophisticated and evasive. Moreover, the emergence of increasingly automated and sophisticated malware and ransomware, powered by Generative AI[170], presents a menacing potential for subverting existing encryption methods[171]. This is primarily due to the immense computational prowess inherent in Generative AI. As these malicious entities persist and advance, they represent a formidable challenge to the cybersecurity landscape, testing the limits of the resilience and robustness of contemporary cybersecurity systems and protocols[172]. The consequences of these developments are far-reaching, with the prospect of malicious AI proving to be devastating to a nation's critical infrastructure, particularly in scenarios involving state-sponsored or malevolent cyber terrorism[173].

**Increased Impersonation and misinformation,** escalation of AI advancements across various domains, visual, speech, audio, and text-based applications, has significantly elevated concerns surrounding personal privacy breaches and impersonation. A pertinent example is the music industry, where AI-driven ghostwriters have released a fake audio tracks emulating the voices of renowned artists like Drake and The Weeknd, both of whom are global music sensations[174]. Tracks like "Heart on My Sleeve" and "Cuff It" featuring AI-rendered versions of Rihanna and Beyoncé's voices[175], have garnered attention for their remarkably convincing mimicry. Consequently, the creative industry faces substantial threats, particularly sectors reliant on advanced artificial intelligence. As reported, these technologies can potentially jeopardize careers within the entertainment industry.

## VI. CONCLUSION

In conclusion, Generative AI opens the door to a world filled with both unprecedented opportunities and inherent risks. Further in-depth research is necessary to comprehend its multifaceted impacts across various sectors better and develop effective mitigation strategies. Striking a balance between the potential benefits and threats posed by Generative AI is essential to serve humanity's needs best. Throughout this paper, we have delved into state-of-the-art models, explored their mathematical foundations, scrutinized their architectural intricacies, and anticipated their evolution in the future. We have also examined prominent tasks, benchmarked state-of-the-art tools against Generative AI, and assessed their real-world applications. The realms of impact, challenges, and future prospects of Generative AI have been thoroughly addressed.

The journey to harness the full potential of Generative AI is ongoing, requiring swift and thoughtful actions from regulatory authorities to ensure order and alignment with the rapid advancements in AI technology sweeping the world. The role of Explainability AI, Responsive AI, and Privacy-Preserving AI becomes increasingly crucial in this context. The future is bright, but as we move forward, maintaining a delicate equilibrium between the opportunities and risks presented by Generative AI is paramount to realizing its full utility and ensuring it serves humanity effectively.

## REFERENCES

[1] J. McCarthy, M. L. Minsky, N. Rochester, I. B. M. Corporation, and C. E. Shannon, "A PROPOSAL FOR THE DARTMOUTH SUMMER RESEARCH PROJECT ON ARTIFICIAL INTELLIGENCE".

[2] C. Zhang and Y. Lu, "Study on artificial intelligence: The state of the art and future prospects," *J. Ind. Inf. Integr.*, vol. 23, p. 100224, Sep. 2021, doi: 10.1016/j.jii.2021.100224.

[3] N. J. Nilsson, *The Quest for Artificial Intelligence*. Cambridge University Press, 2009.

[4] P. Hamet and J. Tremblay, "Artificial intelligence in medicine," *Metabolism*, vol. 69, pp. S36–S40, Apr. 2017, doi: 10.1016/j.metabol.2017.01.011.

[5] R. S. Michalski, J. G. Carbonell, and T. M. Mitchell, *Machine Learning: An Artificial Intelligence Approach*. Springer Science & Business Media, 2013.

[6] D. L. Du Yi, *Artificial Intelligence with Uncertainty*, 2nd ed. Boca Raton: CRC Press, 2016. doi: 10.1201/9781315366951.

[7] F. Rosenblatt, "The perceptron: A probabilistic model for information storage and organization in the brain," *Psychol. Rev.*, vol. 65, no. 6, pp. 386–408, 1958, doi: 10.1037/h0042519.

[8] J. N. Morgan and J. A. Sonquist, "Problems in the Analysis of Survey Data, and a Proposal," *J. Am. Stat. Assoc.*, vol. 58, no. 302, pp. 415–434, Jun. 1963, doi: 10.1080/01621459.1963.10500855.

[9] E. Fix and J. L. Hodges, "Discriminatory Analysis. Nonparametric Discrimination: Consistency Properties," *Int. Stat. Rev. Rev. Int. Stat.*, vol. 57, no. 3, pp. 238–247, 1989, doi: 10.2307/1403797.

[10] J. Platt, "Sequential Minimal Optimization: A Fast Algorithm for Training Support Vector Machines," Apr. 1998, Accessed: Sep. 30, 2023. [Online]. Available: https://www.microsoft.com/en-us/research/publication/sequential-minimal-optimization-a-fast-algorithm-for-training-support-vector-machines/

[11] L. Breiman, "Random Forests," *Mach. Learn.*, vol. 45, no. 1, pp. 5–32, Oct. 2001, doi: 10.1023/A:1010933404324.

[12] Y. Lecun, L. Bottou, Y. Bengio, and P. Haffner, "Gradient-based learning applied to document recognition," *Proc. IEEE*, vol. 86, no. 11, pp. 2278–2324, Nov. 1998, doi: 10.1109/5.726791.

[13] J. J. Hopfield and D. W. Tank, "'Neural' computation of decisions in optimization problems," *Biol. Cybern.*, vol. 52, no. 3, pp. 141–152, Jul. 1985, doi: 10.1007/BF00339943.

[14] S. Hochreiter and J. Schmidhuber, "Long Short-Term Memory," *Neural Comput.*, vol. 9, no. 8, pp. 1735–1780, Nov. 1997, doi: 10.1162/neco.1997.9.8.1735.

[15] M. Schuster and K. K. Paliwal, "Bidirectional recurrent neural networks," *IEEE Trans. Signal Process.*, vol. 45, no. 11, pp. 2673–2681, Nov. 1997, doi: 10.1109/78.650093.

[16] Y. LeCun, Y. Bengio, and G. Hinton, "Deep learning," *Nature*, vol. 521, no. 7553, Art. no. 7553, May 2015, doi: 10.1038/nature14539.

[17] O. Russakovsky *et al.*, "ImageNet Large Scale Visual Recognition Challenge," *Int. J. Comput. Vis.*, vol. 115, no. 3, pp. 211–252, Dec. 2015, doi: 10.1007/s11263-015-0816-y.

[18] S. Cong and Y. Zhou, "A review of convolutional neural network architectures and their optimizations," *Artif. Intell. Rev.*, vol. 56, no. 3, pp. 1905–1969, Mar. 2023, doi: 10.1007/s10462-022-10213-5.

[19] A. Krizhevsky, I. Sutskever, and G. E. Hinton, "ImageNet Classification with Deep Convolutional Neural Networks," in *Advances in Neural Information Processing Systems*, Curran Associates, Inc., 2012. Accessed: Sep. 30, 2023. [Online]. Available: https://proceedings.neurips.cc/paper/2012/hash/c399862d3b9d6b76c8436e924a68c45b-Abstract.html

[20] K. He, X. Zhang, S. Ren, and J. Sun, "Deep Residual Learning for Image Recognition," presented at the Proceedings of the IEEE Conference on Computer Vision and Pattern Recognition, 2016, pp. 770–778. Accessed: Sep. 30, 2023. [Online]. Available: https://openaccess.thecvf.com/content_cvpr_2016/html/He_Deep_Residual_Learning_CVPR_2016_paper.html

[21] G. Huang, Z. Liu, L. van der Maaten, and K. Q. Weinberger, "Densely Connected Convolutional Networks," presented at the Proceedings of the IEEE Conference on Computer Vision and Pattern Recognition, 2017, pp. 4700–4708. Accessed: Sep. 30, 2023. [Online]. Available: https://openaccess.thecvf.com/content_cvpr_2017/html/Huang_Densely_Connected_Convolutional_CVPR_2017_paper.html

[22] A. G. Howard *et al.*, "MobileNets: Efficient Convolutional Neural Networks for Mobile Vision Applications," arXiv.org. Accessed: Sep. 30, 2023. [Online]. Available: https://arxiv.org/abs/1704.04861v1

[23] M. Tan and Q. Le, "EfficientNet: Rethinking Model Scaling for Convolutional Neural Networks," in *Proceedings of the 36th International Conference on Machine Learning*, PMLR, May 2019, pp. 6105–6114. Accessed: Sep. 30, 2023. [Online]. Available: https://proceedings.mlr.press/v97/tan19a.html

[24] A. Vaswani *et al.*, "Attention is All you Need," in *Advances in Neural Information Processing Systems*, Curran Associates, Inc., 2017. Accessed: Aug. 15, 2023. [Online]. Available: https://proceedings.neurips.cc/paper_files/paper/2017/hash/3f5ee243547dee91fbd053c1c4a845aa-Abstract.html

[25] I. J. Goodfellow *et al.*, "Generative Adversarial Networks." arXiv, Jun. 10, 2014. doi: 10.48550/arXiv.1406.2661.

[26] C. Zheng, G. Wu, F. Bao, Y. Cao, C. Li, and J. Zhu, "Revisiting Discriminative vs. Generative Classifiers: Theory and Implications." arXiv, May 29, 2023. doi: 10.48550/arXiv.2302.02334.

[27] E. Brophy, Z. Wang, Q. She, and T. Ward, "Generative Adversarial Networks in Time Series: A Systematic Literature Review," *ACM Comput. Surv.*, vol. 55, no. 10, p. 199:1-199:31, Feb. 2023, doi: 10.1145/3559540.

[28] G. Zhou *et al.*, "Emerging Synergies in Causality and Deep Generative Models: A Survey." arXiv, Sep. 14, 2023. doi: 10.48550/arXiv.2301.12351.

[29] N. R. Mannuru *et al.*, "Artificial intelligence in developing countries: The impact of generative artificial intelligence (AI) technologies for development," *Inf. Dev.*, p. 02666669231200628, Sep. 2023, doi: 10.1177/02666669231200628.

[30] "Introducing ChatGPT." Accessed: Sep. 30, 2023. [Online]. Available: https://openai.com/blog/chatgpt

[31] C. Leiter *et al.*, "ChatGPT: A Meta-Analysis after 2.5 Months." arXiv, Feb. 20, 2023. doi: 10.48550/arXiv.2302.13795.

[32] D. Bank, N. Koenigstein, and R. Giryes, "Autoencoders." arXiv, Apr. 03, 2021. doi: 10.48550/arXiv.2003.05991.

[33] U. Michelucci, "An Introduction to Autoencoders." arXiv, Jan. 11, 2022. doi: 10.48550/arXiv.2201.03898.

[34] J. Zhai, S. Zhang, J. Chen, and Q. He, "Autoencoder and Its Various Variants," in *2018 IEEE International Conference on Systems, Man, and Cybernetics (SMC)*, Oct. 2018, pp. 415–419. doi: 10.1109/SMC.2018.00080.

[35] D. P. Kingma, S. Mohamed, D. Jimenez Rezende, and M. Welling, "Semi-supervised Learning with Deep Generative Models," in *Advances in Neural Information Processing Systems*, Curran Associates, Inc., 2014. Accessed: Aug. 07, 2023. [Online]. Available: https://proceedings.neurips.cc/paper/2014/hash/d523773c6b194f37b938d340d5d02232-Abstract.html

[36] D. P. Kingma and M. Welling, "An Introduction to Variational Autoencoders," *Found. Trends® Mach. Learn.*, vol. 12, no. 4, pp. 307–392, 2019, doi: 10.1561/2200000056.

[37] H. Akrami, A. A. Joshi, J. Li, S. Aydore, and R. M. Leahy, "Brain Lesion Detection Using A Robust Variational Autoencoder and Transfer Learning," in *2020 IEEE 17th International Symposium on Biomedical Imaging (ISBI)*, Apr. 2020, pp. 786–790. doi: 10.1109/ISBI45749.2020.9098405.

[38] X. Shen, B. Liu, Y. Zhou, J. Zhao, and M. Liu, "Remote sensing image captioning via Variational Autoencoder and Reinforcement Learning," *Knowl.-Based Syst.*, vol. 203, p. 105920, Sep. 2020, doi: 10.1016/j.knosys.2020.105920.

[39] G. Zhao and Y. Peng, "Semisupervised SAR image change detection based on a siamese variational autoencoder," *Inf. Process. Manag.*, vol. 59, no. 1, p. 102726, Jan. 2022, doi: 10.1016/j.ipm.2021.102726.

[40] H. W. L. Mak, R. Han, and H. H. F. Yin, "Application of Variational AutoEncoder (VAE) Model and Image Processing Approaches in Game Design," *Sensors*, vol. 23, no. 7, Art. no. 7, Jan. 2023, doi: 10.3390/s23073457.

[41] M. A. Yílmaz, O. Keleş, H. Güven, A. M. Tekalp, J. Malik, and S. Kíranyaz, "Self-Organized Variational Autoencoders (Self-Vae) For Learned Image Compression," in *2021 IEEE International Conference on Image Processing (ICIP)*, Sep. 2021, pp. 3732–3736. doi: 10.1109/ICIP42928.2021.9506041.

[42] Z.-S. Liu, W.-C. Siu, and L.-W. Wang, "Variational AutoEncoder for Reference based Image Super-Resolution," in *2021 IEEE/CVF Conference on Computer Vision and Pattern Recognition Workshops (CVPRW)*, Nashville, TN, USA: IEEE, Jun. 2021, pp. 516–525. doi: 10.1109/CVPRW53098.2021.00063.

[43] G. Carbajal, J. Richter, and T. Gerkmann, "Guided Variational Autoencoder for Speech Enhancement with a Supervised Classifier," in *ICASSP 2021 - 2021 IEEE International Conference on Acoustics, Speech and Signal Processing (ICASSP)*, Jun. 2021, pp. 681–685. doi: 10.1109/ICASSP39728.2021.9414363.

[44] T. Srikotr and K. Mano, "Sub-band Vector Quantized Variational AutoEncoder for Spectral Envelope Quantization," in *TENCON 2019 - 2019 IEEE Region 10 Conference (TENCON)*, Oct. 2019, pp. 296–300. doi: 10.1109/TENCON.2019.8929436.

[45] Z. Wang, "Using Gaussian Process in Clockwork Variational Autoencoder for Video Prediction," in *2022 International Conference on Information Technology Research and Innovation (ICITRI)*, Nov. 2022, pp. 6–11. doi: 10.1109/ICITRI56423.2022.9970241.

[46] M. S. Kim, J. P. Yun, S. Lee, and P. Park, "Unsupervised Anomaly detection of LM Guide Using Variational Autoencoder," in *2019 11th International Symposium on Advanced Topics in Electrical Engineering (ATEE)*, Mar. 2019, pp. 1–5. doi: 10.1109/ATEE.2019.8724998.

[47] C. K. Meher, R. Nayak, and U. C. Pati, "Dual Stream Variational Autoencoder for Video Anomaly Detection in Single Scene Videos," in *2022 2nd Odisha International Conference on Electrical Power Engineering, Communication and Computing Technology (ODICON)*, Nov. 2022, pp. 1–6. doi: 10.1109/ODICON54453.2022.10010086.

[48] H. Yanagihashi and T. Sudo, "Noise-robust Early Detection of Cooling Fan Deterioration with a Variational Autoencoder-based Method," in *2022 9th International Conference on Condition Monitoring and Diagnosis (CMD)*, Nov. 2022, pp. 183–188. doi: 10.23919/CMD54214.2022.9991542.

[49] H. Purohit, T. Endo, M. Yamamoto, and Y. Kawaguchi, "Hierarchical Conditional Variational Autoencoder

Based Acoustic Anomaly Detection," in *2022 30th European Signal Processing Conference (EUSIPCO)*, Aug. 2022, pp. 274–278. doi: 10.23919/EUSIPCO55093.2022.9909785.

[50] B. Shen, L. Yao, and Z. Ge, "Nonlinear probabilistic latent variable regression models for soft sensor application: From shallow to deep structure," *Control Eng. Pract.*, vol. 94, p. 104198, Jan. 2020, doi: 10.1016/j.conengprac.2019.104198.

[51] B. Shen and Z. Ge, "Supervised Nonlinear Dynamic System for Soft Sensor Application Aided by Variational Auto-Encoder," *IEEE Trans. Instrum. Meas.*, vol. 69, no. 9, pp. 6132–6142, Sep. 2020, doi: 10.1109/TIM.2020.2968162.

[52] W. Xie, J. Wang, C. Xing, S. Guo, M. Guo, and L. Zhu, "Variational Autoencoder Bidirectional Long and Short-Term Memory Neural Network Soft-Sensor Model Based on Batch Training Strategy," *IEEE Trans. Ind. Inform.*, vol. 17, no. 8, pp. 5325–5334, Aug. 2021, doi: 10.1109/TII.2020.3025204.

[53] R. Xie, N. M. Jan, K. Hao, L. Chen, and B. Huang, "Supervised Variational Autoencoders for Soft Sensor Modeling With Missing Data," *IEEE Trans. Ind. Inform.*, vol. 16, no. 4, pp. 2820–2828, Apr. 2020, doi: 10.1109/TII.2019.2951622.

[54] F. Guo, R. Xie, and B. Huang, "A deep learning just-in-time modeling approach for soft sensor based on variational autoencoder," *Chemom. Intell. Lab. Syst.*, vol. 197, p. 103922, Feb. 2020, doi: 10.1016/j.chemolab.2019.103922.

[55] L. Li, J. Yan, H. Wang, and Y. Jin, "Anomaly Detection of Time Series With Smoothness-Inducing Sequential Variational Auto-Encoder," *IEEE Trans. Neural Netw. Learn. Syst.*, vol. 32, no. 3, pp. 1177–1191, Mar. 2021, doi: 10.1109/TNNLS.2020.2980749.

[56] N. T. N. Anh, T. Q. Khanh, N. Q. Dat, E. Amouroux, and V. K. Solanki, "Fraud detection via deep neural variational autoencoder oblique random forest," in *2020 IEEE-HYDCON*, Sep. 2020, pp. 1–6. doi: 10.1109/HYDCON48903.2020.9242753.

[57] C. Zhang *et al.*, "A Complete Survey on Generative AI (AIGC): Is ChatGPT from GPT-4 to GPT-5 All You Need?" arXiv, Mar. 21, 2023. doi: 10.48550/arXiv.2303.11717.

[58] K. Han *et al.*, "A Survey on Vision Transformer," *IEEE Trans. Pattern Anal. Mach. Intell.*, vol. 45, no. 1, pp. 87–110, Jan. 2023, doi: 10.1109/TPAMI.2022.3152247.

[59] S. Khan, M. Naseer, M. Hayat, S. W. Zamir, F. S. Khan, and M. Shah, "Transformers in Vision: A Survey," *ACM Comput. Surv.*, vol. 54, no. 10s, p. 200:1-200:41, Sep. 2022, doi: 10.1145/3505244.

[60] T. Lin, Y. Wang, X. Liu, and X. Qiu, "A survey of transformers," *AI Open*, vol. 3, pp. 111–132, Jan. 2022, doi: 10.1016/j.aiopen.2022.10.001.

[61] M. Zong and B. Krishnamachari, "a survey on GPT-3." arXiv, Dec. 01, 2022. Accessed: Aug. 15, 2023. [Online]. Available: http://arxiv.org/abs/2212.00857

[62] B. Ghojogh and A. Ghodsi, "Attention Mechanism, Transformers, BERT, and GPT: Tutorial and Survey," Open Science Framework, preprint, Dec. 2020. doi: 10.31219/osf.io/m6gcn.

[63] C. Zhang *et al.*, "A Complete Survey on Generative AI (AIGC): Is ChatGPT from GPT-4 to GPT-5 All You Need?" arXiv, Mar. 21, 2023. doi: 10.48550/arXiv.2303.11717.

[64] "Generative AI: a game-changer society needs to be ready for," World Economic Forum. Accessed: Aug. 16, 2023. [Online]. Available: https://www.weforum.org/agenda/2023/01/davos23-generative-ai-a-game-changer-industries-and-society-code-developers/

[65] A. Radford, K. Narasimhan, T. Salimans, and I. Sutskever, "Improving Language Understanding by Generative Pre-Training".

[66] I. Solaiman *et al.*, "Release Strategies and the Social Impacts of Language Models".

[67] T. Brown *et al.*, "Language Models are Few-Shot Learners," *Adv. Neural Inf. Process. Syst.*, vol. 33, pp. 1877–1901, 2020.

[68] X. Zheng, C. Zhang, and P. C. Woodland, "Adapting GPT, GPT-2 and BERT Language Models for Speech Recognition," in *2021 IEEE Automatic Speech Recognition and Understanding Workshop (ASRU)*, Dec. 2021, pp. 162–168. doi: 10.1109/ASRU51503.2021.9688232.

[69] A. Shrivastava, R. Pupale, and P. Singh, "Enhancing Aggression Detection using GPT-2 based Data Balancing Technique," in *2021 5th International Conference on Intelligent Computing and Control Systems (ICICCS)*, May 2021, pp. 1345–1350. doi: 10.1109/ICICCS51141.2021.9432283.

[70] M. Tamimi, M. Salehi, and S. Najari, "Deceptive review detection using GAN enhanced by GPT structure and score of reviews," in *2023 28th International Computer Conference, Computer Society of Iran (CSICC)*, Jan. 2023, pp. 1–7. doi: 10.1109/CSICC58665.2023.10105368.

[71] S. Saravanan and K. Sudha, "GPT-3 Powered System for Content Generation and Transformation," in *2022 Fifth International Conference on Computational Intelligence and Communication Technologies (CCICT)*, Jul. 2022, pp. 514–519. doi: 10.1109/CCiCT56684.2022.00096.

[72] K. H. Manodnya and A. Giri, "GPT-K: A GPT-based model for generation of text in Kannada," in *2022 IEEE 4th International Conference on Cybernetics, Cognition and Machine Learning Applications (ICCCMLA)*, Oct. 2022, pp. 534–539. doi: 10.1109/ICCCMLA56841.2022.9989289.

[73] P. Isaranontakul and W. Kreesuradej, "A Study of Using GPT-3 to Generate a Thai Sentiment Analysis of

COVID-19 Tweets Dataset," in *2023 20th International Joint Conference on Computer Science and Software Engineering (JCSSE)*, Jun. 2023, pp. 106–111. doi: 10.1109/JCSSE58229.2023.10201994.

[74] N. Aydın and O. A. Erdem, "A Research On The New Generation Artificial Intelligence Technology Generative Pretraining Transformer 3," in *2022 3rd International Informatics and Software Engineering Conference (IISEC)*, Dec. 2022, pp. 1–6. doi: 10.1109/IISEC56263.2022.9998298.

[75] M. Lajkó, V. Csuvik, and L. Vidács, "Towards JavaScript program repair with Generative Pre-trained Transformer (GPT-2)," in *2022 IEEE/ACM International Workshop on Automated Program Repair (APR)*, May 2022, pp. 61–68. doi: 10.1145/3524459.3527350.

[76] Y. Su, "Computer-generated Humour Based on GPT-2," in *2022 IEEE 2nd International Conference on Data Science and Computer Application (ICDSCA)*, Oct. 2022, pp. 890–893. doi: 10.1109/ICDSCA56264.2022.9987901.

[77] N. Darapaneni, R. Prajeesh, P. Dutta, V. K. Pillai, A. Karak, and A. R. Paduri, "Abstractive Text Summarization Using BERT and GPT-2 Models," in *2023 International Conference on Signal Processing, Computation, Electronics, Power and Telecommunication (IConSCEPT)*, May 2023, pp. 1–6. doi: 10.1109/IConSCEPT57958.2023.10170093.

[78] Y. Liang and Z. Han, "Intelligent Love Letter Generator Based on GPT-2 Model," in *2022 3rd International Conference on Electronic Communication and Artificial Intelligence (IWECAI)*, Jan. 2022, pp. 562–567. doi: 10.1109/IWECAI55315.2022.00115.

[79] D. H. Nguyen, H. L. Pham, and L. Le Thi Trang, "Security of the Cryptosystem GPT Based on Rank Codes and Term-rank Codes," in *2021 International Conference Engineering and Telecommunication (En&T)*, Nov. 2021, pp. 1–5. doi: 10.1109/EnT50460.2021.9681778.

[80] M. Nam, S. Park, and D. S. Kim, "Intrusion Detection Method Using Bi-Directional GPT for in-Vehicle Controller Area Networks," *IEEE Access*, vol. 9, pp. 124931–124944, 2021, doi: 10.1109/ACCESS.2021.3110524.

[81] D. Demırcı, N. şahın, M. şirlancis, and C. Acarturk, "Static Malware Detection Using Stacked BiLSTM and GPT-2," *IEEE Access*, vol. 10, pp. 58488–58502, 2022, doi: 10.1109/ACCESS.2022.3179384.

[82] H. Khan, M. Alam, S. Al-Kuwari, and Y. Faheem, "OFFENSIVE AI: UNIFICATION OF EMAIL GENERATION THROUGH GPT-2 MODEL WITH A GAME-THEORETIC APPROACH FOR SPEAR-PHISHING ATTACKS," in *Competitive Advantage in the Digital Economy (CADE 2021)*, Jun. 2021, pp. 178–184. doi: 10.1049/icp.2021.2422.

[83] H. Liu, Y. Cai, Z. Lin, Z. Ou, Y. Huang, and J. Feng, "Variational Latent-State GPT for Semi-Supervised Task-Oriented Dialog Systems," *IEEEACM Trans. Audio Speech Lang. Process.*, vol. 31, pp. 970–984, 2023, doi: 10.1109/TASLP.2023.3240661.

[84] S. W. Jeong, C. G. Kim, and T. K. Whangbo, "Question Answering System for Healthcare Information based on BERT and GPT," in *2023 Joint International Conference on Digital Arts, Media and Technology with ECTI Northern Section Conference on Electrical, Electronics, Computer and Telecommunications Engineering (ECTI DAMT & NCON)*, Mar. 2023, pp. 348–352. doi: 10.1109/ECTIDAMTNCON57770.2023.10139365.

[85] Y. Zhang, Z. Li, and J. Zhang, "Towards the Use of Pretrained Language Model GPT-2 for Testing the Hypothesis of Communicative Efficiency in the Lexicon," in *2021 International Conference on Asian Language Processing (IALP)*, Dec. 2021, pp. 62–66. doi: 10.1109/IALP54817.2021.9675217.

[86] R. Kinoshita and S. Shiramatsu, "Agent for Recommending Information Relevant to Web-based Discussion by Generating Query Terms using GPT-3," in *2022 IEEE International Conference on Agents (ICA)*, Nov. 2022, pp. 24–29. doi: 10.1109/ICA55837.2022.00011.

[87] C. Treude, "Navigating Complexity in Software Engineering: A Prototype for Comparing GPT-n Solutions," in *2023 IEEE/ACM 5th International Workshop on Bots in Software Engineering (BotSE)*, May 2023, pp. 1–5. doi: 10.1109/BotSE59190.2023.00008.

[88] J. J. Bird, M. Pritchard, A. Fratini, A. Ekárt, and D. R. Faria, "Synthetic Biological Signals Machine-Generated by GPT-2 Improve the Classification of EEG and EMG Through Data Augmentation," *IEEE Robot. Autom. Lett.*, vol. 6, no. 2, pp. 3498–3504, Apr. 2021, doi: 10.1109/LRA.2021.3056355.

[89] P. Maddigan and T. Susnjak, "Chat2VIS: Generating Data Visualizations via Natural Language Using ChatGPT, Codex and GPT-3 Large Language Models," *IEEE Access*, vol. 11, pp. 45181–45193, 2023, doi: 10.1109/ACCESS.2023.3274199.

[90] OpenAI, "GPT-4 Technical Report." arXiv, Mar. 27, 2023. doi: 10.48550/arXiv.2303.08774.

[91] H. Nori, N. King, S. M. McKinney, D. Carignan, and E. Horvitz, "Capabilities of GPT-4 on Medical Challenge Problems." arXiv, Apr. 12, 2023. doi: 10.48550/arXiv.2303.13375.

[92] D. M. Katz, M. J. Bommarito, S. Gao, and P. Arredondo, "GPT-4 Passes the Bar Exam." Rochester, NY, Mar. 15, 2023. doi: 10.2139/ssrn.4389233.

[93] D. Hendrycks *et al.*, "Measuring Massive Multitask Language Understanding," arXiv.org. Accessed: Aug. 22, 2023. [Online]. Available: https://arxiv.org/abs/2009.03300v3

[94] A. Creswell, T. White, V. Dumoulin, K. Arulkumaran, B. Sengupta, and A. A. Bharath, "Generative Adversarial Networks: An Overview," *IEEE Signal*

*Process. Mag.*, vol. 35, no. 1, pp. 53–65, Jan. 2018, doi: 10.1109/MSP.2017.2765202.

[95] A. Aggarwal, M. Mittal, and G. Battineni, "Generative adversarial network: An overview of theory and applications," *Int. J. Inf. Manag. Data Insights*, vol. 1, no. 1, p. 100004, Apr. 2021, doi: 10.1016/j.jjimei.2020.100004.

[96] Z. Zhang, M. Li, and J. Yu, "On the convergence and mode collapse of GAN," in *SIGGRAPH Asia 2018 Technical Briefs*, in SA '18. New York, NY, USA: Association for Computing Machinery, Dec. 2018, pp. 1–4. doi: 10.1145/3283254.3283282.

[97] Bhagyashree, V. Kushwaha, and G. C. Nandi, "Study of Prevention of Mode Collapse in Generative Adversarial Network (GAN)," in *2020 IEEE 4th Conference on Information & Communication Technology (CICT)*, Dec. 2020, pp. 1–6. doi: 10.1109/CICT51604.2020.9312049.

[98] H. Thanh-Tung and T. Tran, "Catastrophic forgetting and mode collapse in GANs," in *2020 International Joint Conference on Neural Networks (IJCNN)*, Jul. 2020, pp. 1–10. doi: 10.1109/IJCNN48605.2020.9207181.

[99] W. Li, L. Fan, Z. Wang, C. Ma, and X. Cui, "Tackling mode collapse in multi-generator GANs with orthogonal vectors," *Pattern Recognit.*, vol. 110, p. 107646, Feb. 2021, doi: 10.1016/j.patcog.2020.107646.

[100] D. Saxena and J. Cao, "Generative Adversarial Networks (GANs): Challenges, Solutions, and Future Directions".

[101] H. Chen, "Challenges and Corresponding Solutions of Generative Adversarial Networks (GANs): A Survey Study," *J. Phys. Conf. Ser.*, vol. 1827, no. 1, p. 012066, Mar. 2021, doi: 10.1088/1742-6596/1827/1/012066.

[102] M. Mirza and S. Osindero, "Conditional Generative Adversarial Nets," arXiv.org. Accessed: Aug. 26, 2023. [Online]. Available: https://arxiv.org/abs/1411.1784v1

[103] G. G. Chrysos, J. Kossaifi, and S. Zafeiriou, "Robust Conditional Generative Adversarial Networks." arXiv, Mar. 13, 2019. doi: 10.48550/arXiv.1805.08657.

[104] A. Radford, L. Metz, and S. Chintala, "Unsupervised Representation Learning with Deep Convolutional Generative Adversarial Networks." arXiv, Jan. 07, 2016. Accessed: Aug. 26, 2023. [Online]. Available: http://arxiv.org/abs/1511.06434

[105] M. Arjovsky, S. Chintala, and L. Bottou, "Wasserstein GAN." arXiv, Dec. 06, 2017. Accessed: Aug. 27, 2023. [Online]. Available: http://arxiv.org/abs/1701.07875

[106] T. C. Koopmans, "Optimum Utilization of the Transportation System," *Econometrica*, vol. 17, pp. 136–146, 1949, doi: 10.2307/1907301.

[107] E. Massart, "Improving weight clipping in Wasserstein GANs," in *2022 26th International Conference on Pattern Recognition (ICPR)*, Montreal, QC, Canada: IEEE, Aug. 2022, pp. 2286–2292. doi: 10.1109/ICPR56361.2022.9956056.

[108] J.-Y. Zhu, T. Park, P. Isola, and A. A. Efros, "Unpaired Image-To-Image Translation Using Cycle-Consistent Adversarial Networks," presented at the Proceedings of the IEEE International Conference on Computer Vision, 2017, pp. 2223–2232. Accessed: Aug. 27, 2023. [Online]. Available: https://openaccess.thecvf.com/content_iccv_2017/html/Zhu_Unpaired_Image-To-Image_Translation_ICCV_2017_paper.html

[109] Y. Choi, M. Choi, M. Kim, J.-W. Ha, S. Kim, and J. Choo, "StarGAN: Unified Generative Adversarial Networks for Multi-Domain Image-to-Image Translation," presented at the Proceedings of the IEEE Conference on Computer Vision and Pattern Recognition, 2018, pp. 8789–8797. Accessed: Aug. 27, 2023. [Online]. Available: https://openaccess.thecvf.com/content_cvpr_2018/html/Choi_StarGAN_Unified_Generative_CVPR_2018_paper.html

[110] T. Karras, T. Aila, S. Laine, and J. Lehtinen, "Progressive Growing of GANs for Improved Quality, Stability, and Variation." arXiv, Feb. 26, 2018. doi: 10.48550/arXiv.1710.10196.

[111] A. Brock, J. Donahue, and K. Simonyan, "Large Scale GAN Training for High Fidelity Natural Image Synthesis." arXiv, Feb. 25, 2019. doi: 10.48550/arXiv.1809.11096.

[112] T. Karras, S. Laine, and T. Aila, "A Style-Based Generator Architecture for Generative Adversarial Networks," presented at the Proceedings of the IEEE/CVF Conference on Computer Vision and Pattern Recognition, 2019, pp. 4401–4410. Accessed: Aug. 29, 2023. [Online]. Available: https://openaccess.thecvf.com/content_CVPR_2019/html/Karras_A_Style-Based_Generator_Architecture_for_Generative_Adversarial_Networks_CVPR_2019_paper.html

[113] T. Karras, S. Laine, M. Aittala, J. Hellsten, J. Lehtinen, and T. Aila, "Analyzing and Improving the Image Quality of StyleGAN," presented at the Proceedings of the IEEE/CVF Conference on Computer Vision and Pattern Recognition, 2020, pp. 8110–8119. Accessed: Aug. 29, 2023. [Online]. Available: https://openaccess.thecvf.com/content_CVPR_2020/html/Karras_Analyzing_and_Improving_the_Image_Quality_of_StyleGAN_CVPR_2020_paper.html

[114] X. Chen, Y. Duan, R. Houthooft, J. Schulman, I. Sutskever, and P. Abbeel, "InfoGAN: Interpretable Representation Learning by Information Maximizing Generative Adversarial Nets," in *Advances in Neural Information Processing Systems*, Curran Associates, Inc., 2016. Accessed: Aug. 29, 2023. [Online]. Available: https://proceedings.neurips.cc/paper_files/paper/2016/

hash/7c9d0b1f96aebd7b5eca8c3edaa19ebb-Abstract.html

[115] T. Salimans *et al.*, "Improved Techniques for Training GANs," in *Advances in Neural Information Processing Systems*, Curran Associates, Inc., 2016. Accessed: Aug. 29, 2023. [Online]. Available: https://proceedings.neurips.cc/paper_files/paper/2016/hash/8a3363abe792db2d8761d6403605aeb7-Abstract.html

[116] J. Donahue, P. Krähenbühl, and T. Darrell, "Adversarial Feature Learning." arXiv, Apr. 03, 2017. doi: 10.48550/arXiv.1605.09782.

[117] A. Ramesh, P. Dhariwal, A. Nichol, C. Chu, and M. Chen, "Hierarchical Text-Conditional Image Generation with CLIP Latents," arXiv.org. Accessed: Oct. 06, 2023. [Online]. Available: https://arxiv.org/abs/2204.06125v1

[118] J. Ho *et al.*, "IMAGEN VIDEO: HIGH DEFINITION VIDEO GENERATION WITH DIFFUSION MODELS".

[119] U. Singer *et al.*, "Make-A-Video: Text-to-Video Generation without Text-Video Data." arXiv, Sep. 29, 2022. doi: 10.48550/arXiv.2209.14792.

[120] Y. Li *et al.*, "Competition-level code generation with AlphaCode," *Science*, vol. 378, no. 6624, pp. 1092–1097, Dec. 2022, doi: 10.1126/science.abq1158.

[121] "StarCoder: A State-of-the-Art LLM for Code." Accessed: Oct. 06, 2023. [Online]. Available: https://huggingface.co/blog/starcoder

[122] M. Chen *et al.*, "Evaluating Large Language Models Trained on Code." arXiv, Jul. 14, 2021. doi: 10.48550/arXiv.2107.03374.

[123] "A systematic review of artificial intelligence-based music generation: Scope, applications, and future trends - ScienceDirect." Accessed: Oct. 07, 2023. [Online]. Available: https://www.sciencedirect.com/science/article/pii/S0957417422013537

[124] "MuseNet." Accessed: Oct. 07, 2023. [Online]. Available: https://openai.com/research/musenet

[125] P. Dhariwal, H. Jun, C. Payne, J. W. Kim, A. Radford, and I. Sutskever, "Jukebox: A Generative Model for Music." arXiv, Apr. 30, 2020. doi: 10.48550/arXiv.2005.00341.

[126] "How To Give Your Living Room a Luxurious Makeover," House Beautiful. Accessed: Sep. 26, 2023. [Online]. Available: https://www.housebeautiful.com/room-decorating/living-family-rooms/g715/designer-living-rooms/

[127] "File:Bowie-state-university-science-building.jpg - Wikipedia." Accessed: Sep. 26, 2023. [Online]. Available: https://commons.wikimedia.org/wiki/File:Bowie-state-university-science-building.jpg

[128] N. Kaur and P. Singh, "Conventional and contemporary approaches used in text to speech synthesis: a review," *Artif. Intell. Rev.*, vol. 56, no. 7, pp. 5837–5880, Jul. 2023, doi: 10.1007/s10462-022-10315-0.

[129] A. Wali *et al.*, "Generative adversarial networks for speech processing: A review," *Comput. Speech Lang.*, vol. 72, p. 101308, Mar. 2022, doi: 10.1016/j.csl.2021.101308.

[130] J. A. Rodriguez, D. Vazquez, I. Laradji, M. Pedersoli, and P. Rodriguez, "FigGen: Text to Scientific Figure Generation." arXiv, Jun. 21, 2023. Accessed: Oct. 07, 2023. [Online]. Available: http://arxiv.org/abs/2306.00800

[131] "Minerva: Solving Quantitative Reasoning Problems with Language Models." Accessed: Oct. 07, 2023. [Online]. Available: https://blog.research.google/2022/06/minerva-solving-quantitative-reasoning.html

[132] R. Taylor *et al.*, "Galactica: A Large Language Model for Science." arXiv, Nov. 16, 2022. doi: 10.48550/arXiv.2211.09085.

[133] C. Plut and P. Pasquier, "Generative music in video games: State of the art, challenges, and prospects," *Entertain. Comput.*, vol. 33, p. 100337, Mar. 2020, doi: 10.1016/j.entcom.2019.100337.

[134] S. Wang *et al.*, "ReelFramer: Co-creating News Reels on Social Media with Generative AI." arXiv, Apr. 19, 2023. Accessed: Oct. 11, 2023. [Online]. Available: http://arxiv.org/abs/2304.09653

[135] T. H. D. and R. Bean, "The Impact of Generative AI on Hollywood and Entertainment," MIT Sloan Management Review. Accessed: Oct. 11, 2023. [Online]. Available: https://sloanreview.mit.edu/article/the-impact-of-generative-ai-on-hollywood-and-entertainment/

[136] "Runway AI: Tech Behind Everything Everywhere All At Once." Accessed: Oct. 11, 2023. [Online]. Available: https://topten.ai/ai-tech-behind-everything-everywhere-all-at-once/

[137] L. Rai, C. Deng, and F. Liu, "Developing Massive Open Online Course Style Assessments using Generative AI Tools," in *2023 IEEE 6th International Conference on Electronic Information and Communication Technology (ICEICT)*, Jul. 2023, pp. 1292–1294. doi: 10.1109/ICEICT57916.2023.10244824.

[138] J. Qadir, "Engineering Education in the Era of ChatGPT: Promise and Pitfalls of Generative AI for Education," in *2023 IEEE Global Engineering Education Conference (EDUCON)*, May 2023, pp. 1–9. doi: 10.1109/EDUCON54358.2023.10125121.

[139] W. M. Lim, A. Gunasekara, J. L. Pallant, J. I. Pallant, and E. Pechenkina, "Generative AI and the future of education: Ragnarök or reformation? A paradoxical perspective from management educators," *Int. J. Manag. Educ.*, vol. 21, no. 2, p. 100790, Jul. 2023, doi: 10.1016/j.ijme.2023.100790.

[140] Y.-X. Li and N.-C. Tai, "Teaching at the Right Moment: A Generative AI-Enabled Bedtime Storybook Generation System Communicating Timely Issues," in *2023 International Conference on Consumer Electronics - Taiwan (ICCE-Taiwan)*, Jul. 2023, pp. 157–158. doi: 10.1109/ICCE-Taiwan58799.2023.10226626.

[141] E. A. Alasadi and C. R. Baiz, "Generative AI in Education and Research: Opportunities, Concerns, and Solutions," *J. Chem. Educ.*, vol. 100, no. 8, pp. 2965–2971, Aug. 2023, doi: 10.1021/acs.jchemed.3c00323.

[142] H. Yu, Z. Liu, and Y. Guo, "Application Status, Problems and Future Prospects of Generative AI in Education," in *2023 5th International Conference on Computer Science and Technologies in Education (CSTE)*, Apr. 2023, pp. 1–7. doi: 10.1109/CSTE59648.2023.00065.

[143] M. Kuzlu, Z. Xiao, S. Sarp, F. O. Catak, N. Gurler, and O. Guler, "The Rise of Generative Artificial Intelligence in Healthcare," in *2023 12th Mediterranean Conference on Embedded Computing (MECO)*, Jun. 2023, pp. 1–4. doi: 10.1109/MECO58584.2023.10155107.

[144] A. Jadon and S. Kumar, "Leveraging Generative AI Models for Synthetic Data Generation in Healthcare: Balancing Research and Privacy," in *2023 International Conference on Smart Applications, Communications and Networking (SmartNets)*, Jul. 2023, pp. 1–4. doi: 10.1109/SmartNets58706.2023.10215825.

[145] Z. Shen, F. Ding, A. Jolfaei, K. Yadav, S. Vashisht, and K. Yu, "DeformableGAN: Generating Medical Images With Improved Integrity for Healthcare Cyber Physical Systems," *IEEE Trans. Netw. Sci. Eng.*, vol. 10, no. 5, pp. 2584–2596, Sep. 2023, doi: 10.1109/TNSE.2022.3190765.

[146] "AWS Announces AWS HealthScribe, a New Generative AI-Powered Service that Automatically Creates Clinical Documentation," Press Center. Accessed: Oct. 11, 2023. [Online]. Available: https://press.aboutamazon.com/2023/7/aws-announces-aws-healthscribe-a-new-generative-ai-powered-service-that-automatically-creates-clinical-documentation

[147] B. Tang, J. Ewalt, and H.-L. Ng, "Generative AI Models for Drug Discovery," in *Biophysical and Computational Tools in Drug Discovery*, A. K. Saxena, Ed., in Topics in Medicinal Chemistry. , Cham: Springer International Publishing, 2021, pp. 221–243. doi: 10.1007/7355_2021_124.

[148] W. P. Walters and M. Murcko, "Assessing the impact of generative AI on medicinal chemistry," *Nat. Biotechnol.*, vol. 38, no. 2, Art. no. 2, Feb. 2020, doi: 10.1038/s41587-020-0418-2.

[149] X. Zeng *et al.*, "Deep generative molecular design reshapes drug discovery," *Cell Rep. Med.*, vol. 3, no. 12, p. 100794, Dec. 2022, doi: 10.1016/j.xcrm.2022.100794.

[150] A. Madani *et al.*, "Large language models generate functional protein sequences across diverse families," *Nat. Biotechnol.*, vol. 41, no. 8, Art. no. 8, Aug. 2023, doi: 10.1038/s41587-022-01618-2.

[151] E. Sevgen *et al.*, "ProT-VAE: Protein Transformer Variational AutoEncoder for Functional Protein Design." bioRxiv, p. 2023.01.23.525232, Jan. 24, 2023. doi: 10.1101/2023.01.23.525232.

[152] "Cognizant and Google Cloud Expand Alliance to Bring AI to Enterprise Clients," News | Cognizant Technology Solutions. Accessed: Oct. 11, 2023. [Online]. Available: https://news.cognizant.com/2023-05-09-Cognizant-and-Google-Cloud-Expand-Alliance-to-Bring-AI-to-Enterprise-Clients

[153] "Generative AI to Become a $1.3 Trillion Market by 2032, Research Finds | Press | Bloomberg LP," *Bloomberg L.P.* Accessed: Oct. 12, 2023. [Online]. Available: https://www.bloomberg.com/company/press/generative-ai-to-become-a-1-3-trillion-market-by-2032-research-finds/

[154] T. H. Baek, "Digital Advertising in the Age of Generative AI," *J. Curr. Issues Res. Advert.*, vol. 44, no. 3, pp. 249–251, Jul. 2023, doi: 10.1080/10641734.2023.2243496.

[155] J. Huh, M. R. Nelson, and C. A. Russell, "ChatGPT, AI Advertising, and Advertising Research and Education," *J. Advert.*, vol. 52, no. 4, pp. 477–482, Aug. 2023, doi: 10.1080/00913367.2023.2227013.

[156] J. Ford, V. Jain, K. Wadhwani, and D. G. Gupta, "AI advertising: An overview and guidelines," *J. Bus. Res.*, vol. 166, p. 114124, Nov. 2023, doi: 10.1016/j.jbusres.2023.114124.

[157] A. Beheshti *et al.*, "ProcessGPT: Transforming Business Process Management with Generative Artificial Intelligence," in *2023 IEEE International Conference on Web Services (ICWS)*, Jul. 2023, pp. 731–739. doi: 10.1109/ICWS60048.2023.00099.

[158] "Amazon launches generative AI to help sellers write product descriptions," US About Amazon. Accessed: Oct. 12, 2023. [Online]. Available: https://www.aboutamazon.com/news/small-business/amazon-sellers-generative-ai-tool

[159] "ChatGPT In Healthcare: What Science Says," The Medical Futurist. Accessed: Aug. 24, 2023. [Online]. Available: https://medicalfuturist.com/chatgpt-in-healthcare-what-the-science-says/

[160] "Generative AI and the future of work in America | McKinsey." Accessed: Oct. 10, 2023. [Online]. Available: https://www.mckinsey.com/mgi/our-research/generative-ai-and-the-future-of-work-in-america#/

[161] A. Zarifhonarvar, "Economics of ChatGPT: A Labor Market View on the Occupational Impact of Artificial Intelligence." Rochester, NY, Feb. 07, 2023. doi: 10.2139/ssrn.4350925.

[162] "Future of Work Report: AI at Work." Accessed: Oct. 14, 2023. [Online]. Available: https://economicgraph.linkedin.com/research/future-of-work-report-ai

[163] "Jobs of Tomorrow: Large Language Models and Jobs," World Economic Forum. Accessed: Oct. 14, 2023. [Online]. Available: https://www.weforum.org/whitepapers/jobs-of-tomorrow-large-language-models-and-jobs/

[164] "Automation or augmentation? This is how AI will be integrated into the jobs of tomorrow," World Economic Forum. Accessed: Oct. 14, 2023. [Online]. Available: https://www.weforum.org/agenda/2023/09/ai-automation-augmentation-workplace-jobs-of-tomorrow/

[165] "We always hear that AI will take our jobs. But what jobs will it create?," World Economic Forum. Accessed: Oct. 14, 2023. [Online]. Available: https://www.weforum.org/agenda/2023/09/jobs-ai-will-create/

[166] K. Michael, R. Abbas, and G. Roussos, "AI in Cybersecurity: The Paradox," *IEEE Trans. Technol. Soc.*, vol. 4, no. 2, pp. 104–109, Jun. 2023, doi: 10.1109/TTS.2023.3280109.

[167] S. Oh and T. Shon, "Cybersecurity Issues in Generative AI," in *2023 International Conference on Platform Technology and Service (PlatCon)*, Aug. 2023, pp. 97–100. doi: 10.1109/PlatCon60102.2023.10255179.

[168] M. Gupta, C. Akiri, K. Aryal, E. Parker, and L. Praharaj, "From ChatGPT to ThreatGPT: Impact of Generative AI in Cybersecurity and Privacy," *IEEE Access*, vol. 11, pp. 80218–80245, 2023, doi: 10.1109/ACCESS.2023.3300381.

[169] "AI-Based Cybercrime Tools WormGPT and FraudGPT Could Be The Tip of the Iceberg | SlashNext," SlashNext |. Accessed: Aug. 24, 2023. [Online]. Available: https://slashnext.com/blog/ai-based-cybercrime-tools-wormgpt-and-fraudgpt-could-be-the-tip-of-the-iceberg/

[170] I.-C. Mihai, "The Transformative Impact of Artificial Intelligence on Cybersecurity," *Int. J. Inf. Secur. Cybercrime*, vol. 12, p. 9, 2023.

[171] P. V. Falade, "Decoding the Threat Landscape : ChatGPT, FraudGPT, and WormGPT in Social Engineering Attacks," *Int. J. Sci. Res. Comput. Sci. Eng. Inf. Technol.*, pp. 185–198, Oct. 2023, doi: 10.32628/CSEIT2390533.

[172] M. Mozes, X. He, B. Kleinberg, and L. D. Griffin, "Use of LLMs for Illicit Purposes: Threats, Prevention Measures, and Vulnerabilities." arXiv, Aug. 24, 2023. doi: 10.48550/arXiv.2308.12833.

[173] "Pause Giant AI Experiments: An Open Letter," Future of Life Institute. Accessed: Aug. 24, 2023. [Online]. Available: https://futureoflife.org/open-letter/pause-giant-ai-experiments/

[174] J. Coscarelli, "An A.I. Hit of Fake 'Drake' and 'The Weeknd' Rattles the Music World," *The New York Times*, Apr. 19, 2023. Accessed: Oct. 11, 2023. [Online]. Available: https://www.nytimes.com/2023/04/19/arts/music/ai-drake-the-weeknd-fake.html

[175] B. Lane, "An AI-generated Rihanna cover of Beyoncé's 'Cuff It' is going viral, and it could open up a new legal nightmare for the music industry," Insider. Accessed: Oct. 14, 2023. [Online]. Available: https://www.insider.com/rihanna-ai-cuff-it-cover-legal-nightmare-music-industry-2023-4